# Penalty Constraints and Kernelization of M-Estimation Based Fuzzy C—Means

Jingwei Liu[a,*]    Meizhi Xu[b]

[a] *School of Mathematics and System Sciences, Beihang University,Beijing,100191, P.R China*
[b] *Department of Mathematics, Tsinghua University, Beijing, 100084, P.R China*

**Abstract:** A framework of M-estimation based fuzzy C–means clustering (MFCM) algorithm is proposed with iterative reweighted least squares (IRLS) algorithm, and penalty constraint and kernelization extensions of MFCM algorithms are also developed. Introducing penalty information to the object functions of MFCM algorithms, the spatially constrained fuzzy c-means (SFCM) is extended to penalty constraints MFCM algorithms (*abbr.* pMFCM). Substituting the Euclidean distance with kernel method, the MFCM and pMFCM algorithms are extended to kernelized MFCM (*abbr.* KMFCM) and kernelized pMFCM (*abbr.* pKMFCM) algorithms. The performances of MFCM, pMFCM, KMFCM and pKMFCM algorithms are evaluated in three tasks: pattern recognition on 10 standard data sets from UCI Machine Learning databases, noise image segmentation performances on a synthetic image, a magnetic resonance brain image (MRI), and image segmentation of a standard images from Berkeley Segmentation Dataset and Benchmark. The experimental results demonstrate the effectiveness of our proposed algorithms in pattern recognition and image segmentation.

## 1. Introduction

Fuzzy C-means (FCM) algorithm is one of the most popular fuzzy clustering method widely used in various tasks of pattern recognition, data mining, image processing, gene expression data recognition, *etc.* [1,2,3,4]. Modifying and generalizing the FCMalgorithm is a prevailing research stream in fuzzy clustering in recent decades. Many solutions have been developed to modify FCM to improve its robustness and classification accuracy. Based on the modification modes, the solutions can be grossly dividedinto three categories : modifying the objective function[2,3,5,6,7,8,9], kernelizing the inner–product norm [2,9,10,11,12,13], and introducing the spatial penalty [14,15,16,17,18,19,20]. Furthermore, kernelizing the inner–product norm can also be treated as a mode of modifying the objective function of FCM. And, the kernelizing SFCM (SKFCM) is equivalent to spatially constraining kernelized FCM (KFCM).

---

[*] Corresponding author. *Email address*: jwliu@buaa.edu.cn.    Nov. 18, 2012

To modify the objective function of standard FCM, an M-estimation [21,22] method from robust statistics is introduced. Two introduction styles are investigated in FCM extension. The first one is with the original idea of M-estimation [22] in robust statistics [2,5,11]. The second one is using the residual fitting problem with IRLS algorithm [6,8] [23,24]. In [6], a class of attribute C-means clustering (AMC) algorithm is proposed to generalize FCM by introducing a concept of stable function, where stable function is the weight function in M-estimation, [23,24]. Four kinds of stable functions ( squared stable function, Cauchy stable function, general $l_p$ stable function and exponential stable function) are investigated in [6,25]. And, [8] generalize the AMC to Bezdek type AMC (called FAMC), and kernelize FAMC models in [12].

As there are many types of weight functions in M-estimation, and extending FCM with spatial constraints and kernelization could provide flexibility and robustness of FCM [10,11,15,20] especially in image segmentation. The main motivation of this paper is to develop the general framework of FCM based on M-estimation with IRLS method (*called* MFCM), and extend it to spatial constrained penalty and kernelization models. We incorporate two kinds of penalty information into the membership functions of theMFCM algorithms [17,19], and propose a class of penalty constrained MFCM (*called* pMFCM) algorithms. For each penalty function, we investigate two kinds of neighborhood information choices[26]. Since kernel method is a popular approach in pattern recognition, signal processing, *etc.* [27,28] , the pMFCM algorithms are finally extended to kernelized pMFCM (*abbr.* pKMFCM).

To evaluate the performance of pMFCM and pKMFCM algorithms, we demonstrate the classification accuracy on 10 standard UCI data sets [8], and the image segmentation capability on a synthetic image, a standard synthetic magnetic resonance image (MRI) [10,11,13], and two standard images from Berkeley image segmentation data sets [29]. The classification and segmentation results demonstrate the effectiveness of our proposed models.

The rest of the paper is organized as follows: In section 2, a brief review of FCM and KFCM is given. We prove that the possible choices kernel function are uncountable and propose MFCM algorithms. In section 3, pMFCM with two kinds of penalty informationare developed. In section 4, KMFCM and pKMFCM algorithms are proposed. In section 5, the classification and image segmentation experiments are investigated. Finally, the paper concludes in the last section.

## 2. M-estimation based FCM and Kernelization

### *2.1. Brief review of FCM and KFCM*

The mathematical foundation of FCM is to minimize the following least-squares objective function,

$$J_m(U, v) = \sum_{k=1}^{C} \sum_{n=1}^{N} u_{kn}^m \| x_n - v_k \|^2 \tag{1}$$

constrained by

$$U \in M_{fc} = \{ U \in R^{C \times N} \mid u_{kn} \in [0,1], \sum_{k=1}^{C} u_{kn} = 1, \forall n, k.\} \tag{2}$$

where, $\{x_1, x_2, \cdots, x_n\} \in \Omega \subseteq R^d$ are $n$ samples in $d$-dimensional pattern space, $v = \{v_1, v_2, \cdots, v_C\}$ are $C$ cluster centers or centroids, $u_{kn}$ is the membership function of the $n$th sample belonging to the $k$th centroid. $m > 1$ is a weighting exponent or fuzziness index to control the "fuzziness" in the objective function. The distance $\|\cdot\|$ between $x_n$ and the centroid $v_k$ is Euclidean distance or Mahalanobis distance [1,9]. We only discuss Euclidean distance in this paper.

To obtain high classification accuracy, the samples are mapped into high dimensional space with nonlinear mapping function $y = \Phi(x)$, satisfying

$$\Phi : x \in \Omega \subseteq R^d \mapsto \Phi(x) \in H \subseteq R^N, \tag{3}$$

where $d \leq N \leq \infty$. The inner–product in high dimension space called a kernel function

$$\kappa(x, z) = <\Phi(x), \Phi(z)>, \forall x, z \in \Omega \tag{4}$$

Four widely used basic kernel function in kernel analysis are as follows [30,31,32,33],

1) Linear : $\kappa(x, z) = <x, z> = x^T z$
2) Polynomial : $\kappa(x, z) = (\beta x^T z + \theta)^d, \beta > 0, \theta > 0, d \in N^+$.
3) radial basic function (RBF) : $\kappa(x, z) = \exp(-\beta \| x - z \|^2), \beta > 0$.
4) sigmoid : $\kappa(x, z) = \tanh(\beta x^T z + \theta), \beta > 0, \theta > 0$.

However, the following proposition shows that more complicated kernels can be created by simple kernels [27].

**Proposition 1.** Let $\kappa_1$ and $\kappa_2$ be kernels over $\Omega \times \Omega$, $\Omega \subseteq R^d$, $a \in R^+$, $f(\cdot)$ a real–valued function on $\Omega$, $\Phi : \Omega \to R^N$ with $\kappa_3$ a kernel over $R^N \times R^N$, and $\mathbf{B}$ a symmetric positive semi–definite $d \times d$ matrix, $p(\cdot)$ is a polynomial with positive coefficients. Then the following functions are kernels:

$$\begin{aligned}
&1) \quad \kappa(x,z) = \kappa_1(x,z) + \kappa_2(x,z) \quad 2) \quad \kappa(x,z) = a\kappa_1(x,z) \\
&3) \quad \kappa(x,z) = \kappa_1(x,z)\kappa_2(x,z) \quad 4) \quad \kappa(x,z) = f(x)f(z) \\
&5) \quad \kappa(x,z) = \kappa_3(\Phi(x), \Phi(z)) \quad 6) \quad \kappa(x,z) = x^T \mathbf{B} z \\
&7) \quad \kappa(x,z) = p(\kappa_1(x,z)) \quad 8) \quad \kappa(x,z) = \exp(\kappa_1(x,z))
\end{aligned} \tag{5}$$

Since

$$\|\Phi(x_n)-\Phi(v_k)\|^2 = <\Phi(x_n)-\Phi(v_k),\Phi(x_n)-\Phi(v_k)>$$
$$= <\Phi(x_n),\Phi(x_n)> + <\Phi(v_k),\Phi(v_k)> -2<\Phi(x_n),\Phi(v_k)> \qquad (6)$$
$$= \kappa(x_n,x_n)+\kappa(v_k,v_k)-2\kappa(x_n,v_k)$$

There are two ways to kernelize FCM in fuzzy clustering. The first KFCM algorithm (*called* KFCM-F) is constructed by minimizing the objective function as follows

$$J_{Ker}(U,v) = \sum_{k=1}^{C}\sum_{n=1}^{N} u_{kn}^m \|\Phi(x_n)-\Phi(v_k)\|^2$$
$$= \sum_{k=1}^{C}\sum_{n=1}^{N} u_{kn}^m (\kappa(x_n,x_n)+\kappa(v_k,v_k)-2\kappa(x_n,v_k)) \qquad (7)$$

Zhou *et al.* [34] and Graves, *et al.* [9] extend the polynomial KFCM based on minimizing the following object function

$$P(U,v) = \sum_{k=1}^{C}\sum_{n=1}^{N} u_{kn}^m \|\Phi(x_n)-v_k^*\|^2 \qquad (8)$$

where $v_k^*$ is the centroid in kernel space, it is so called the KFCM–K algorithm. In this paper, we only address the KFCM-F algorithm and still denote it KFCM.

Chen, *et al.*[13] discuss the multiple-kernel FCM algorithm based on Proposition 1. In fact, there are infinite combinations of kernels, and the conclusion can be obtained from the following Theorem 1 [35,36].

**Theorem 1.** Let $J$ denote set of all kernel algorithms,

$$J = \{J \mid J = \sum_{k=1}^{C}\sum_{n=1}^{N} u_{kn}^m (\kappa(x_n,x_n)+\kappa(v_k,v_k)-2\kappa(x_n,v_k)), \forall \kappa: \mathbf{R}^d \times \mathbf{R}^d \mapsto \mathbf{R}\}. \qquad (9)$$

Then,

$$card(J) = \aleph,$$

where $card(J)$ is the cardinal number of set $J$, and $\aleph$ denotes uncountable.

**Proof.** According to the formula 4) and 6) in Proposition 1, and

$$card(\{f \mid \forall f: R^d \to R\}) = \aleph, \quad card(\{B \mid \forall B \in R^{d \times d}\}) = \aleph.$$

Then, $card(J) > card(\{f \mid \forall f: R^d \to R\}) + card(\{B \mid \forall B \in R^{d \times d}\}) = \aleph.$ ∎

Theorem 1 shows that there are uncountable kinds of KFCM algorithms with different kernels, which leads to finding concise and effective KFCM an important research field in fuzzy clustering, our scheme is to develop FCM based on M-estimation of robust statistics and extend it to kernelization form.

## 2.2. MFCM algorithm

M-Estimation is the maximum likelihood type estimations (MLEs) proposed by Huber [21,22], which is the extension of MLE. For $x_1, x_2, \cdots, x_n \stackrel{iid}{\sim} F(x-\theta)$, where $\theta$ is a location parameter, M-estimation is to maximize the objective function

$$\min_{\theta} \sum_{i=1}^{n} \rho(x_i - \theta) \tag{10}$$

where, $\rho$ is an arbitrary function of symmetric convex function increasing less rapidly than the square. The original motivation of M-estimation is for the robust estimation of linear regression [23,24,37], suppose $r_i$ is the $i-$th residual of $i-$th sample data and its fit value, and the so called M-estimator is to estimate parameter $\theta = (p_1, p_2, \cdots, p_s)$ from $\min \sum_{i=1}^{n} \rho(r_i)$.

$$\min \sum_{i=1}^{n} \rho(r_i) \tag{11}$$

Calculating first derivation of $\sum_{i=1}^{n} \rho(r_i)$, the solution is

$$\sum_{i=1}^{n} \psi(r_i) \frac{\partial r_i}{\partial p_j} = \sum_{i=1}^{n} w(r_i) r_i \frac{\partial r_i}{\partial p_j} = 0, j = 1, 2, \cdots, s. \tag{12}$$

The above solution equals to solve the following IRLS problem [24,25],

$$\min \sum_{i=1}^{n} w(r_i^{(k-1)}) r_i^2 \tag{13}$$

where $k$ is the iteration times, and $w(r_i^{(k-1)})$ in (12) is the same function $w(r_i)$ in (11) at the $k$ th iteration. The theoretical difference between formula (11) and formula (13) lies in that formula(11) is based on M-estimation while formula(13) is an iterative system, The common ground is they have same solution formula (12) in the sense of iterative approximation. [2,5,10,11,23,38,39,40] focus on formula (11)(12), while [6,8,12] carry on research based on formula (13)(12).

Cheng [6,25] explains this theory based on the relationship between $\rho(x)$ and $w(x)$ as follows:

$$\rho(x) = \int_{0}^{x} 2sw(s)ds.$$

And, $\psi(x,\theta) = \frac{\partial \rho(x,\theta)}{\partial \theta}$ is called an influence function, $w(x) = \frac{\psi(x)}{x}$ is called a weight function.[1]

---

[1] Cheng [6] defines $w(x) = \frac{\psi(x)}{2x}$ as stable function, and $\rho(x)$ satisfies the conditions that

1) $\rho(x)$ is a positive differential function in $[0, \infty)$.

2) $w(x) = \frac{\rho'(x)}{2x}$ is a positive non-increasable function.

Seven common types of robust function $\rho(x)$ and weight function $w(x)$ are as follows [6,23,25] ($\beta > 0$),

1) $L_2$ : $\rho(x) = \dfrac{x^2}{2}$, $w(x) = 1$. [1]

2) $L_1 - L_2$ : $\rho(x) = 2((1+\dfrac{x^2}{2})^{\frac{1}{2}} - 1)$, $w(x) = (1+\dfrac{x^2}{2})^{-\frac{1}{2}}$.

3) Huber : $\rho(x) = \begin{cases} \dfrac{x^2}{2}, & |x| \le \beta \\ \beta(|x| - \dfrac{\beta}{2}), & |x| > \beta \end{cases}$ $w(x) = \begin{cases} 1, & |x| \le \beta \\ \dfrac{\beta}{|x|}, & |x| > \beta \end{cases}$. [2]

4) German–Maclure : $\rho(x) = \dfrac{x^2}{2(1+x^2)}$, $w(x) = \dfrac{1}{(1+x^2)^2}$.

5) Welsch : $\rho(x) = \dfrac{\beta^2}{2}(1 - \exp(-(\dfrac{x}{\beta})^2))$, $w(x) = \exp(-(\dfrac{x}{\beta})^2)$. [3]

6) Cauchy : $\rho(x) = \dfrac{\beta^2}{2}\log(1+(\dfrac{x}{\beta})^2)$, $w(x) = (1+(\dfrac{x}{\beta})^2)^{-1}$. [4]

7) Fair : $\rho(x) = \beta^2[\dfrac{|x|}{\beta} - \log(1+\dfrac{|x|}{\beta})]$, $w(x) = (1+\dfrac{|x|}{\beta})$.

Based on the M-estimation discussion, the alternative FCM (AFCM) [2] and KFCM [11] are obtained by modifying the Euclidean norm of FCM with RBF kernel norm. This kind of extension could be expressed by

$$P(U,v) = \sum_{k=1}^{C} \sum_{n=1}^{N} u_{kn}^m \rho(\|x_n - v_k\|^2) \qquad (14)$$

where $\rho(\|x_n - v_k\|)$ takes the RBF kernel norm. And, Frigui, *et al.* [38], Wang [39] and Winkler, *et al.* [40] call this kind of KFCM as robust FCM (RFCM). Generally, $\rho(\|x_n - v_k\|)$ is

---

[1] $L_2$ is the special case of $L_p$ : $\rho(x) = \dfrac{|x|^p}{p}$, $w(x) = |x|^{p-2}$, $0 < p \le 2$. And $L_2$ is equivalent to the squared stable function in [6].

[2] If we modify the general $l_p$ [6,25] robust function as follows,

1) $l_p (0 < p \le 2)$ : $\rho(x) = \begin{cases} \dfrac{x^2}{2}, & |x| \le \beta \\ \dfrac{\beta^{2-p}}{p}(|x|^p - \dfrac{2-p}{2}\beta^p), & |x| > \beta \end{cases}$, $w(x) = \begin{cases} 1, & |x| \le \beta \\ (\dfrac{\beta}{|x|})^{2-p}, & |x| > \beta \end{cases}$.

2) $l_p (p = 0)$ : $\rho(x) = \begin{cases} \dfrac{x^2}{2}, & |x| \le \beta \\ \dfrac{\beta^2}{2}(1+2\ln(\dfrac{x}{\beta})), & |x| > \beta \end{cases}$, $w(x) = \begin{cases} 1, & |x| \le \beta \\ (\dfrac{\beta}{|x|})^2, & |x| > \beta \end{cases}$.

Huber robust function is the $l_1$ robust function.

[3] Welsch robust function is equivalent to exponential stable function in Cheng (1993,1998).

[4] Cauchy robust function is equivalent to the Cauchy stable function in [6,25].

not a Bezdek type FCM, theextension of KFCM based on robust function $\rho$ also faces the limitation.

Cheng [6] and Liu,*et al.* [8] propose another FCM extension way according to M-estimation based on the IRLS algorithm, called attribute C-means clustering (AMC) and Bezdek type fuzzy AMC (FAMC) respectively, which is an iterative algorithm to minimize the following objective function,

$$P(U,v) = \sum_{k=1}^{C} \sum_{n=1}^{N} \rho(u_{kn}^{\frac{m}{2}} \|x_n - v_k\|) \tag{15}$$

According to IRLS algorithm, it is equal to iterate the following objective function,

$$\min \sum_{k=1}^{C} \sum_{n=1}^{N} w(R_{kn}^{(i)}) u_{kn}^{m} \|x_n - v_k\|^2 \tag{16}$$

where $R_{kn} = u_{kn}^{\frac{m}{2}} \|x_n - v_k\|$, $w(x)$ is the weight function and denote $R_{kn}^{(i)}$ as the $i$ iterative time value of $R_{kn}$. We call formula (15) M-estimation based FCM (MFCM). It can be treated as a kind of weighted FCM, where the weight functions are derived from M-estimation theory.

## 3. Penalty constrained MFCM

Two types of penalty constrained MFCM with membership function are incorporated to the MFCM algorithms in this section, which are denoted as pMFCM S-I algorithm and pMFCM S-II algorithm .

### 3.1. pMFCM S-I algorithms

To incorporate spatial information to MFCM, the following objective with restriction of membership function [15] is proposed as :

$$J_I(U,v) = \sum_{k=1}^{C} \sum_{n=1}^{N} w_{kn} u_{kn}^{m} d_{kn}^{2} + \frac{\gamma}{2} \sum_{k=1}^{C} \sum_{n=1}^{N} w_{kn} u_{kn}^{m} \sum_{j \in N_n} \sum_{l \in M_k} u_{jl}^{m} \tag{17}$$

where $M_k = \{1,2,\cdots,C\} \setminus \{k\}$, $N_n$ is the set of neighbors of $x_n$, $m > 1$. Two kinds of neighbors [26] are investigated in this paper, first–order neighborhood and second–order neighborhood, denoted as I–order (Fig.1(a)) and II–order (Fig.1(b)(c)) respectively, and two kinds of II–order neighborhoods are denoted as nn-I (Fig.1(b)) and nn-II (Fig.1(c)) respectively.

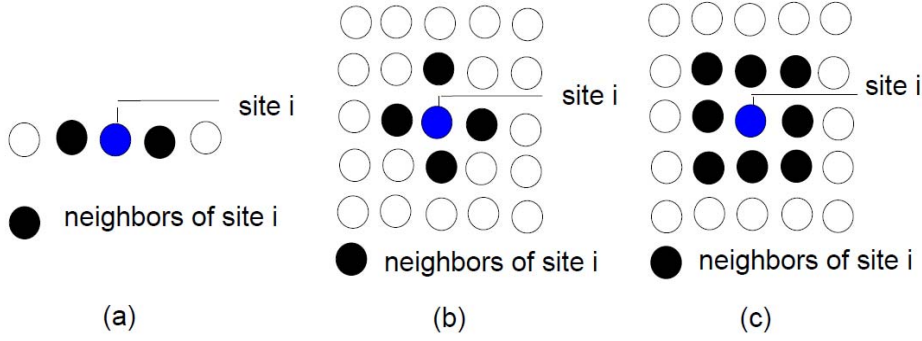

Figure 1: (a) first–order neighborhood (I–order). (b) second–order neighborhood (nn–I). (c) second–order neighborhood (nn–II).

To calculate the $(U, v)$ parameters minimizing formula (13), we give the following updating scheme. Denote

$$J_I^{(i)}(U^{(i)}, v) = Q^{(i)}(U^{(i)}, v) + \frac{\gamma}{2} \sum_{k=1}^{C} \sum_{n=1}^{N} w_{kn}^{(i)} (u_{kn}^{(i)})^m \sum_{j \in N_n} \sum_{l \in M_k} (u_{jl}^{(i)})^m.$$

$$J_I^{(i)}(U, v^{(i+1)}) = Q^{(i)}(U, v^{(i+1)}) + \frac{\gamma}{2} \sum_{k=1}^{C} \sum_{n=1}^{N} w_{kn}^{(i)} u_{kn}^m \sum_{j \in N_n} \sum_{l \in M_k} u_{jl}^m.$$

**Theorem 2.** Fix $m \in (1, \infty)$, let $\Omega = \{x_1, \cdots, x_N\}$ be sample set with $C$ centroids, where $x_n$ is $d$–dimensional vector ($d \geq 1$). Define

$$I_n^{(i+1)} = \{k \mid 1 \leq k \leq C, d_{kn}^{(i+1)} = \| x_n - v_k^{(i+1)} \| = 0\}.$$
$$\widetilde{I_n^{(i+1)}} = \{1, 2, \cdots, C\} - I_n^{(i+1)}. \tag{18}$$

Then $(U, v)$ may be globally minimal for $J_I(U, v)$ by updating $J_I^{(i)}(U^{(i)}, v)$ and $J_I^{(i)}(U, v^{(i+1)})$ only if

$$v_k^{(i+1)} = \frac{\sum_{n=1}^{N} w_{kn}^{(i)} (u_{kn}^{(i)})^m x_n}{\sum_{n=1}^{N} w_{kn}^{(i)} (u_{kn}^{(i)})^m} \tag{19}$$

And,

$$I_n^{(i+1)} = \varnothing \Rightarrow u_{kn}^{(i+1)} = \frac{\left( w_{kn}^{(i)} ((d_{kn}^{(i+1)})^2 + \gamma \sum_{j \in N_n} \sum_{l \in M_k} (u_{jl}^{(i)})^m ) \right)^{-\frac{1}{m-1}}}{\sum_{k=1}^{C} \left( w_{kn}^{(i)} ((d_{kn}^{(i+1)})^2 + \gamma \sum_{j \in N_n} \sum_{l \in M_k} (u_{jl}^{(i)})^m ) \right)^{-\frac{1}{m-1}}}$$

or  \hfill (20)

$$I_n^{(i+1)} \neq \varnothing \Rightarrow u_{kn}^{(i+1)} = 0, \forall k \in \widetilde{I_n^{(i+1)}} \text{ and } \sum_{k \in I_n^{(i+1)}} u_{kn}^{(i+1)} = 1.$$

**Proof.** Since the penalty function does not depend on $v_k$, the iteration of $v_k^{(i+1)}$ is calculated by minimizing $J_I^{(i)}(U^{(i)}, v)$ with partial derivative about $v$, and we obtain formula (19).

Utilizing the Lagrange multiplier [15], the iteration of $u_{kn}^{(i+1)}$ is calculated by minimizing $J_I^{(i)}(U, v^{(i+1)})$ with constrains of $\sum_{k=1}^{C} u_{kn} = 1$. Calculating the partial derivative with respect to $u_{kn}$, we obtain that when $I_n^{(i+1)} = \varnothing$,

$$\frac{\partial}{\partial u_{kn}} \left( \sum_{k=1}^{C} \sum_{n=1}^{N} w_{kn}^{(i)} u_{kn}^m ((d_{kn}^{(i+1)})^2 + \frac{\gamma}{2} \sum_{j \in N_n} \sum_{l \in M_k} (u_{jl}^{(i)})^m) + \sum_{n=1}^{N} \lambda_n (1 - \sum_{k=1}^{C} u_{kn}) \right) \quad (21)$$

$$= m w_{kn}^{(i)} u_{kn}^{m-1} \left( (d_{kn}^{(i+1)})^2 + \gamma \sum_{j \in N_n} \sum_{l \in M_k} (u_{jl}^{(i)})^m \right) - \lambda_n$$

where the factor $\frac{1}{2}$ of $\gamma$ vanished since the derivative operator works in a term corresponding to the product of $u_{kn}$ and its neighbors, and a reverse product term of its neighbors and $u_{kn}$ [15]. Let formula (21) be zero, we obtain

$$u_{kn} = \left( \frac{m w_{kn}^{(i)} ((d_{kn}^{(i+1)})^2 + \gamma \sum_{j \in N_n} \sum_{l \in M_k} (u_{jl}^{(i)})^m)}{\lambda_n} \right)^{-\frac{1}{m-1}}$$

Employing the constraint equation $\sum_{k=1}^{C} u_{kn} = 1$ to above formula, we obtain

$$\sum_{k=1}^{C} \left( \frac{m w_{kn}^{(i)} ((d_{kn}^{(i+1)})^2 + \gamma \sum_{j \in N_n} \sum_{l \in M_k} (u_{jl}^{(i)})^m)}{\lambda_n} \right)^{-\frac{1}{m-1}} = 1, \quad (22)$$

which leads to

$$\lambda_n^{-\frac{1}{m-1}} = \sum_{k=1}^{C} \left( m w_{kn}^{(i)} ((d_{kn}^{(i+1)})^2 + \gamma \sum_{j \in N_n} \sum_{l \in M_k} (u_{jl}^{(i)})^m) \right)^{-\frac{1}{m-1}}. \quad (23)$$

Combining formulae (22)(23), we obtain

$$u_{kn} = \frac{\left( w_{kn}^{(i)} ((d_{kn}^{(i+1)})^2 + \gamma \sum_{j \in N_n} \sum_{l \in M_k} (u_{jl}^{(i)})^m) \right)^{-\frac{1}{m-1}}}{\sum_{k=1}^{C} \left( w_{kn}^{(i)} ((d_{kn}^{(i+1)})^2 + \gamma \sum_{j \in N_n} \sum_{l \in M_k} (u_{jl}^{(i)})^m) \right)^{-\frac{1}{m-1}}}.$$

Denote $u_{kn}$ as $u_{kn}^{(i+1)}$, we obtain update formula (20).

When $I_n^{(i+1)} \neq \varnothing$, $u_{kn}^{(i+1)} = 0, \forall k \in \widetilde{I_n^{(i+1)}}$. Constrained to $\sum_{k=1}^{C} u_{kn}^{(i+1)} = 1$, we obtain that

$$\sum_{k \in I_n^{(i+1)}} u_{kn}^{(i+1)} = 1. \qquad \blacksquare$$

According to the limitation theorem of mathematical analysis, it is true that

$$\lim_{i \to \infty} J_I^{(i)}(U^{(i)}, v) = \min_U J_I(U, v), \quad \lim_{i \to \infty} J_I^{(i)}(U, v^{(i+1)}) = \min_v J_I(U, v),$$

and

$$\lim_{i \to \infty} (U^{(i)}, v^{(i)}) = \arg \min_{(U, v)} J_I(U, v).$$

Hence, we can obtain the update solution of formula (17). The model of formula (17) with update formulae(19)(20) is called pMFCM S-I algorithm. Since FCM is a special case of MFCM with $L_2$ weight function, pMFCM S-I with $L_2$ weight function is SFCM in [15].

### *3.2. pMFCM S-II algorithms*

To incorporate spatial information to MFCM algorithm, another restriction of membership function [10] is proposed as:

$$J_{II}(U, v) = \sum_{k=1}^{C} \sum_{n=1}^{N} w_{kn} u_{kn}^m d_{kn}^2 + \frac{\gamma}{N_R} \sum_{k=1}^{C} \sum_{n=1}^{N} w_{kn} u_{kn}^m \sum_{j \in N_n} (1 - u_{kj})^m. \qquad (24)$$

where $m > 1$, $N_n$ is the set of neighbors existing in a window around $x_n$ ($x_n$ is excluded), and $N_R$ is the element number of $N_n$ (see Fig.1), where we set $N_R = 2$ in Fig1(a), $N_R = 4$ in Fig1(b) and $N_R = 8$ in Fig1(c) respectively. If $x_n$ is a sequence, Fig1(a) takes the special case of both Fig1(b) and Fig1(c). We denote the Fig1(b) and Fig1(c) penalty information as "nn-I" and "nn-II" respectively.

To minimize the $J_{II}(U, v)$, we adopt the same updating method discussed in Section 3.1. Define,

$$J_{II}^{(i)}(U^{(i)}, v) = Q^{(i)}(U^{(i)}, v) + \frac{\gamma}{N_R} \sum_{k=1}^{C} \sum_{n=1}^{N} w_{kn}^{(i)} (u_{kn}^{(i)})^m \sum_{j \in N_n} (1 - u_{kj}^{(i)})^m.$$

$$J_{II}^{(i)}(U, v^{(i+1)}) = Q^{(i)}(U, v^{(i+1)}) + \frac{\gamma}{N_R} \sum_{k=1}^{C} \sum_{n=1}^{N} w_{kn}^{(i)} u_{kn}^m \sum_{j \in N_n} (1 - u_{kj})^m.$$

And, the updating theorem is described as follows.

**Theorem 3.** Fix $m \in (1, \infty)$, let $\Omega = \{x_1, \cdots, x_N\}$ be sample set with $C$ centroids, where $x_n$

is $d$ –dimensional vector ($d \geq 1$). Define

$$I_n^{(i+1)} = \{k \mid 1 \leq k \leq C, d_{kn}^{(i+1)} = \|x_n - v_k^{(i+1)}\| = 0\}.$$
$$\widetilde{I_n^{(i+1)}} = \{1, 2, \cdots, C\} - I_n^{(i+1)}. \tag{25}$$

Then $(U, v)$ may be globally minimal for $J_{II}(U, v)$ by updating $J_{II}^{(i)}(U^{(i)}, v)$ and $J_{II}^{(i)}(U, v^{(i+1)})$ only if

$$v_k^{(i+1)} = \frac{\sum_{n=1}^{N} w_{kn}^{(i)} (u_{kn}^{(i)})^m x_n}{\sum_{n=1}^{N} w_{kn}^{(i)} (u_{kn}^{(i)})^m} \tag{26}$$

And,

$$I_n^{(i+1)} = \varnothing \Rightarrow u_{kn}^{(i+1)} = \frac{\left( w_{kn}^{(i)} ((d_{kn}^{(i+1)})^2 + \frac{\gamma}{N_R} \sum_{j \in N_n} (1 - u_{kj}^{(i)})^m) \right)^{-\frac{1}{m-1}}}{\sum_{k=1}^{C} \left( w_{kn}^{(i)} ((d_{kn}^{(i+1)})^2 + \frac{\gamma}{N_R} \sum_{j \in N_n} (1 - u_{kj}^{(i)})^m) \right)^{-\frac{1}{m-1}}}$$

or $\tag{27}$

$$I_n^{(i+1)} \neq \varnothing \Rightarrow u_{kn}^{(i+1)} = 0, \forall k \in \widetilde{I_n^{(i+1)}} \text{ and } \sum_{k \in I_n^{(i+1)}} u_{kn}^{(i+1)} = 1.$$

**Proof.** Similar to the proof in [10], since the penalty function does not depend on $v_k$, the iteration of $v_k^{(i+1)}$ is obtained by calculating the partial derivative of $J_{II}^{(i)}(U^{(i)}, v)$ with respect to $v$, and we obtain formula (26).

When $I_n^{(i+1)} = \varnothing$, utilizing the Lagrange multiplier as in [10], the iteration of $u_{kn}^{(i+1)}$ is calculated by minimizing $J_{II}^{(i)}(U, v^{(i+1)})$ with constrains of $\sum_{k=1}^{C} u_{kn} = 1$. Calculating the partial derivative with respect to $u_{kn}$, we obtain

$$\frac{\partial}{\partial u_{kn}} \left( \sum_{k=1}^{C} \sum_{n=1}^{N} w_{kn}^{(i)} u_{kn}^m ((d_{kn}^{(i+1)})^2 + \frac{\gamma}{N_R} \sum_{j \in N_n} (1 - u_{kj}^{(i)})^m) + \sum_{n=1}^{N} \lambda_n (1 - \sum_{k=1}^{C} u_{kn}) \right)$$
$$= m w_{kn}^{(i)} u_{kn}^{m-1} \left( (d_{kn}^{(i+1)})^2 + \frac{\gamma}{N_R} \sum_{j \in N_n} (1 - u_{kj}^{(i)})^m \right) - \lambda_n \tag{28}$$

Let the above formula be zero, we obtain

$$u_{kn} = \left( \frac{m w_{kn}^{(i)} ((d_{kn}^{(i+1)})^2 + \frac{\gamma}{N_R} \sum_{j \in N_n} (1 - u_{kj}^{(i)})^m)}{\lambda_n} \right)^{-\frac{1}{m-1}}$$

Employing the constraint equation $\sum_{k=1}^{C} u_{kn} = 1$, we obtain

$$\sum_{k=1}^{C} \left( \frac{mw_{kn}^{(i)}((d_{kn}^{(i+1)})^2 + \frac{\gamma}{N_R} \sum_{j \in N_n} (1-u_{kj}^{(i)})^m)}{\lambda_n} \right)^{-\frac{1}{m-1}} = 1 \tag{29}$$

which leads to

$$\lambda_n^{-\frac{1}{m-1}} = \sum_{k=1}^{C} \left( mw_{kn}^{(i)}((d_{kn}^{(i+1)})^2 + \frac{\gamma}{N_R} \sum_{j \in N_n} (1-u_{kj}^{(i)})^m) \right)^{-\frac{1}{m-1}} \tag{30}$$

Combining formulae (29)(30), we obtain

$$u_{kn}^{(i+1)} u_{kn} = \frac{\left( w_{kn}^{(i)}((d_{kn}^{(i+1)})^2 + \frac{\gamma}{N_R} \sum_{j \in N_n} (1-u_{kj}^{(i)})^m) \right)^{-\frac{1}{m-1}}}{\sum_{k=1}^{C} \left( w_{kn}^{(i)}((d_{kn}^{(i+1)})^2 + \frac{\gamma}{N_R} \sum_{j \in N_n} (1-u_{kj}^{(i)})^m) \right)^{-\frac{1}{m-1}}}$$

When $I_n^{(i+1)} \neq \varnothing$, $u_{kn}^{(i+1)} = 0, \forall k \in \widetilde{I_n^{(i+1)}}$. Constrained to $\sum_{k=1}^{C} u_{kn}^{(i+1)} = 1$, we obtain that $\sum_{k \in I_n^{(i+1)}} u_{kn}^{(i+1)} = 1$. ∎

According to the limit theorem of mathematical analysis, we can obtain that

$$\lim_{i \longrightarrow \infty} (U^{(i)}, v^{(i)}) = \arg \min_{(U,v)} J_{II}(U, v).$$

Therefore, the optimization solution of formula(24) is obtained by Theorem 3. This kind of pMFCM algorithm with formulae(24)(26)(27) is called as pMFCM S-II algorithm.

When the penalty factor $\gamma = 0$, both pMFCM S-I and pMFCM S-II are MFCM algorithm.

## 4. KMFCM and pKMFCM

### *4.1. Brief review of KFCM–F*

Kernel technique is a popular method in pattern recognition and machine learning. We will extend the pMFCM to kernelized pMFCM in kernel space. As discussed in Section 2.1, there are uncountable choices of kernel function. As linear kernel is the special case of Polynomial kernel, we only discuss the Polynomial, RBF and Tanh kernel functions [9].

There are two kinds of kernelization modeling for FCM, KFCM–F and KFCM–K [9], where KFCM–F denotes that the prototypes constructed in the feature space, while KFCM–K denotes that the prototypes are developed in the kernel space and inversely mapped to feature space to obtain the prototypes of feature space. We only discuss the KFCM–F type kernelization of MFCM and pMFCM in this paper.

The KFCM–F algorithm modifies the objective function of FCM as follows,

$$J_m(U,v) = \sum_{k=1}^{C} \sum_{n=1}^{N} u_{kn}^m \| \Phi(x_n) - \Phi(v_k) \|^2. \quad (31)$$

MFCM algorithms can be easily extend to KMFCM by substituting the Euclidean norm with different kernel norm by minimizing the following objective function,

$$P(U,v) = \sum_{k=1}^{C} \sum_{n=1}^{N} \rho(u_{kn}^{\frac{m}{2}} \| \Phi(x_n) - \Phi(v_k) \|) \quad (32)$$

It is equal to iteratively minimize the following objective function according to IRLS algorithm,

$$Q_{Ker}^{(i)}(U,v) = \sum_{k=1}^{C} \sum_{n=1}^{N} w(R_{kn}^{(i)}) u_{kn}^m \| \Phi(x_n) - \Phi(v_k) \|^2 \quad (33)$$

where, $R_{kn}^{(i)}$ denote the $i$ iterative value of $R_{kn} = u_{kn}^{\frac{m}{2}} \| \Phi(x_n) - \Phi(v_k) \|$, $w(x)$ is a weight function. We call the kernelized MFCM as KMFCM.

Theoretically, pMFCM S-I algorithms and pMFCM S-II algorithms can be obtained by substituting the Euclidean norm with different kernel norm, we called them pKMFCM S-I algorithms and pKMFCM S-II algorithms respectively.

### 4.2. pKMFCM S-I

The pKMFCM S-I algorithm is to minimize the following objective function with IRLS algorithm

$$J_{KerI}(U,v) = \sum_{k=1}^{C} \sum_{n=1}^{N} w_{kn}^{(i)} u_{kn}^m \| \Phi(x_n) - \Phi(v_k) \|^2 + \frac{\gamma}{2} \sum_{k=1}^{C} \sum_{n=1}^{N} w_{kn}^{(i)} u_{kn}^m \sum_{j \in N_n} \sum_{l \in M_k} u_{jl}^m. \quad (34)$$

where $N_n$ and $M_k$ are defined as in section 3.1. $w_{kn}^{(i)}$ is the $i$ th iteration value of $w(u_{kn}^{\frac{m}{2}} \| \Phi(x_n) - \Phi(v_k) \|)$.

To calculate the $(U,v)$ parameters minimizing $J_{KerI}(U,v)$, we give the following updating scheme. Denote

$$J^{(i)}_{KerI}(U^{(i)}, v) = \sum_{k=1}^{C}\sum_{n=1}^{N} w^{(i)}_{kn} u^{m}_{kn} \|\Phi(x_n) - \Phi(v_k)\|^2 + \frac{\gamma}{2}\sum_{k=1}^{C}\sum_{n=1}^{N} w^{(i)}_{kn}(u^{(i)}_{kn})^m \sum_{j \in N_n}\sum_{l \in M_k} s_{l}(u^{(i)}_{jl})^m.$$

$$J^{(i)}_{KerI}(U, v^{(i+1)}) = \sum_{k=1}^{C}\sum_{n=1}^{N} w^{(i)}_{kn} u^{m}_{kn} \|\Phi(x_n) - \Phi(v^{(i+1)}_k)\|^2 + \frac{\gamma}{2}\sum_{k=1}^{C}\sum_{n=1}^{N} w^{(i)}_{kn} u^{m}_{kn} \sum_{j \in N_n}\sum_{l \in M_k} u^{m}_{jl}.$$

**Theorem 4.** Fix $m \in (1, \infty)$, let $\Omega = \{x_1, \cdots, x_N\}$ be sample set with $C$ centroids, where $x_n$ is $d$–dimensional vector ($d \geq 1$). Define

$$I^{(i+1)}_n = \{k \mid 1 \leq k \leq C, D^{(i+1)}_{kn} = \|\Phi(x_n) - \Phi(v^{(i+1)}_k)\| = 0\}.$$
$$\widetilde{I^{(i+1)}_n} = \{1, 2, \cdots, C\} - I^{(i+1)}_n. \tag{35}$$

Then $(U, v)$ of pKMFCM S-I may be globally minimal for $J_{KerI}(U, v)$ by updating $J^{(i)}_{KerI}(U^{(i)}, v)$ and $J^{(i)}_{KerI}(U, v^{(i+1)})$ with RBF, Poly and Tanh kernels only if

$$RBF : v^{(i+1)}_k = \frac{\sum_{n=1}^{N} w^{(i)}_{kn}(u^{(i)}_{kn})^m K(x_n, v_k) x_n}{\sum_{n=1}^{N} w^{(i)}_{kn}(u^{(i)}_{kn})^m K(x_n, v_k)}$$

$$Poly : v^{(i+1)}_k = \frac{\sum_{n=1}^{N} w^{(i)}_{kn}(u^{(i)}_{kn})^m (K(x_n, v_k))^{\frac{d-1}{d}} x_n}{\sum_{n=1}^{N} w^{(i)}_{kn}(u^{(i)}_{kn})^m (K(v_k, v_k))^{\frac{d-1}{d}}} \tag{36}$$

$$Tanh : v^{(i+1)}_k = \frac{\sum_{n=1}^{N} w^{(i)}_{kn}(u^{(i)}_{kn})^m (1 - K^2(x_n, v_k)) x_n}{\sum_{n=1}^{N} w^{(i)}_{kn}(u^{(i)}_{kn})^m (1 - K^2(v_k, v_k))}$$

And,

$$I^{(i+1)}_n = \varnothing \Rightarrow u^{(i+1)}_{kn} = \frac{\left(w^{(i)}_{kn}(\|\Phi(x_n) - \Phi(v^{(i+1)}_k)\|^2 + \gamma \sum_{j \in N_n}\sum_{l \in M_k}(u^{(i)}_{jl})^m)\right)^{-\frac{1}{m-1}}}{\sum_{k=1}^{C}\left(w^{(i)}_{kn}(\|\Phi(x_n) - \Phi(v^{(i+1)}_k)\|^2 + \gamma \sum_{j \in N_n}\sum_{l \in M_k}(u^{(i)}_{jl})^m)\right)^{-\frac{1}{m-1}}}$$

or $\tag{37}$

$$I^{(i+1)}_n \neq \varnothing \Rightarrow u^{(i+1)}_{kn} = 0, \forall k \in \widetilde{I^{(i+1)}_n} \text{ and } \sum_{k \in I^{(i+1)}_n} u^{(i+1)}_{kn} = 1.$$

## *4.3. pKMFCM S-II*

The pKMFCM S-II algorithm is to minimize the following objective function with IRLS

algorithm

$$J_{KerII}(U,v) = \sum_{k=1}^{C}\sum_{n=1}^{N} w_{kn}^{(i)} u_{kn}^m \| \Phi(x_n) - \Phi(v_k) \|^2 \frac{\gamma}{N_R} \sum_{k=1}^{C}\sum_{n=1}^{N} w_{kn}^{(i)} u_{kn}^m \sum_{j \in N_n} (1-u_{kj})^m. \quad (38)$$

where $N_n$ is defined as in section 3.2. $w_{kn}^{(i)}$ is the $i$ th iteration value of $w(u_{kn}^{\frac{m}{2}} \| \Phi(x_n) - \Phi(v_k) \|)$.

To calculate the $(U,v)$ parameters minimizing $J_{KerII}(U,v)$, we give the following updating scheme. Denote

$$J_{KerII}^{(i)}(U^{(i)}, v) = \sum_{k=1}^{C}\sum_{n=1}^{N} w_{kn}^{(i)} u_{kn}^m \| \Phi(x_n) - \Phi(v_k) \|^2 + \frac{\gamma}{N_R} \sum_{k=1}^{C}\sum_{n=1}^{N} w_{kn}^{(i)} (u_{kn}^{(i)})^m \sum_{j \in N_n} (1-u_{kj}^{(i)})^m.$$

$$J_{KerII}^{(i)}(U, v^{(i+1)}) = \sum_{k=1}^{C}\sum_{n=1}^{N} w_{kn}^{(i)} u_{kn}^m \| \Phi(x_n) - \Phi(v_k^{(i+1)}) \|^2 + \frac{\gamma}{N_R} \sum_{k=1}^{C}\sum_{n=1}^{N} w_{kn}^{(i)} u_{kn}^m \sum_{j \in N_n} (1-u_{kj})^m.$$

**Theorem 5.** Fix $m \in (1,\infty)$, let $\Omega = \{x_1, \cdots, x_N\}$ be sample set with $C$ centroids, where $x_n$ is $d$ –dimensional vector ($d \geq 1$). Define

$$\begin{aligned}I_n^{(i+1)} &= \{k \mid 1 \leq k \leq C, D_{kn}^{(i+1)} = \| \Phi(x_n) - \Phi(v_k^{(i+1)}) \| = 0\}, \\ \widetilde{I_n^{(i+1)}} &= \{1, 2, \cdots, C\} - I_n^{(i+1)}.\end{aligned} \quad (39)$$

Then $(U,v)$ may be globally minimal for $J_{KerII}(U,v)$ by updating $J_{KerII}^{(i)}(U^{(i)}, v)$ and $J_{KerII}^{(i)}(U, v^{(i+1)})$ only if

$$\begin{aligned} RBF: v_k^{(i+1)} &= \frac{\sum_{n=1}^{N} w_{kn}^{(i)} (u_{kn}^{(i)})^m K(x_n, v_k) x_n}{\sum_{n=1}^{N} w_{kn}^{(i)} (u_{kn}^{(i)})^m K(x_n, v_k)} \\ Poly: v_k^{(i+1)} &= \frac{\sum_{n=1}^{N} w_{kn}^{(i)} (u_{kn}^{(i)})^m (K(x_n, v_k))^{\frac{d-1}{d}} x_n}{\sum_{n=1}^{N} w_{kn}^{(i)} (u_{kn}^{(i)})^m (K(v_k, v_k))^{\frac{d-1}{d}}} \\ Tanh: v_k^{(i+1)} &= \frac{\sum_{n=1}^{N} w_{kn}^{(i)} (u_{kn}^{(i)})^m (1-K^2(x_n, v_k)) x_n}{\sum_{n=1}^{N} w_{kn}^{(i)} (u_{kn}^{(i)})^m (1-K^2(v_k, v_k))} \end{aligned} \quad (40)$$

And,

$$I_n^{(i+1)} = \emptyset \Rightarrow u_{kn}^{(i+1)} = \frac{\left( w_{kn}^{(i)} (\|\Phi(x_n) - \Phi(v_k^{(i+1)})\|^2 + \gamma \sum_{j \in N_n} (1-u_{kj})^m) \right)^{-\frac{1}{m-1}}}{\sum_{k=1}^{C} \left( w_{kn}^{(i)} (\|\Phi(x_n) - \Phi(v_k^{(i+1)})\|^2 + \gamma \sum_{j \in N_n} (1-u_{kj})^m) \right)^{-\frac{1}{m-1}}}$$

or (41)

$$I_n^{(i+1)} \neq \emptyset \Rightarrow u_{kn}^{(i+1)} = 0, \forall k \in \widetilde{I_n^{(i+1)}} \text{ and } \sum_{k \in I_n^{(i+1)}} u_{kn}^{(i+1)} = 1.$$

The proof of Theorem 4 and Theorem 5 are similar to Theorem 2 and Theorem 3. When $\gamma = 0$, the $(U, v)$ update formulae(36)(37)(40)(41) in pKMFCM S-I and pKMFCM S-II degenerate to the parameter estimation formulae of KMFCM.

## *4.4. Uniform updating of the pKMFCM*

Since, KMFCM and pMFCM are a special case of pKMFCM, and MFCM is the special case of both KMFCM and pMFCM. We give a uniform updating algorithm. Given a weight function and a kernel function, the different of pKMFCM S-I algorithm and pKMFCM S-II algorithm are only different in the penalty functions. We describe the updating procedures of the pKMFCM S-I and pKMFCM S-II algorithms in a uniform algorithm framework as follows,

**Algorithm 1. pKMFCM S-I & S-II**
1) Fix $m > 1, \gamma \geq 0$, fix the weight function $w(x)$, and fix the kernel function $K(\cdot,\cdot)$ (RBF, Poly, Tanh). Randomly initialize $U^{(0)} = (u_{kn}^{(0)})_{C \times N}$, $W^{(0)} = (w_{kn}^{(0)})_{C \times N}$ in $[0,1]$. Set maximum updating times $T$.
2) For pKMFCM S-I, calculating centroid $v_k^{(1)}$ using formula (36). Compute membership functions $u_{kn}^{(1)}$ using formula (37). For pMFCM S-II, calculating centroid $v_k^{(1)}$ using formula (40). Compute membership functions $u_{kn}^{(1)}$ using formula (41). Then calculate $W^{(1)}$ according to weight function $w(x)$.
3) For $i = i+1$, For pMFCM S-I, update $v_k^{(i)}$, $u_{kn}^{(i)}$ using formula (36) (37) respectively. For pMFCM S-II, update $v_k^{(i)}$, $u_{kn}^{(i)}$ using formula (40) (41) respectively. And update $w_{kn}^{(i)}$ according to weight function $w(x)$.
4) If $|U^{(i+1)} - U^{(i)}| < \varepsilon$ or $i > T$, stop updating. Or, go to step 3) and repeat until convergence.

The optimization of penalty factor $\gamma$ is described as follows. For each investigated data set $\Omega$, we split the data set $\Omega$ to a validation set $\Omega_v$ and a training data $\Omega_c$. The pKMFCM algorithms are trained on $\Omega_c$ and tested on $\Omega_v$. We choose the recommended parameter values according the performance on validation set $\Omega_v$. The evaluation criterion is defined as the cross–validation error as follows:

$$E_{Kerv} = \sum_{k=1}^{C} \sum_{n \in \Omega_v} w_{kn} u_{kn}^m d_{kn}^2.$$

where, $m > 1$, $d_{kn} = \| \Phi(x_n) - \Phi(v_k) \|^2$, $w_{kn} = w(u_{kn}^{\frac{m}{2}} d_{kn})$. The optimization of $\gamma$ schemes in pKMFCM S-I and pKMFCM S-II are described as follows:

**Algorithm 2. Optimization of $\gamma$ in pKMFCM algorithms**

1) Fix weight function $w(x)$, and kernel function $K(\cdot,\cdot)$ (RBF, Poly, Tanh). For pKMFCM, Apply KMFCM (or pKMFCM with $\gamma = 0$) to the sample space $\Omega$. And, set $\gamma = 1$ and maximum updating times $T_\gamma$.

2) For pKMFCM S-I, update $\gamma = 0.1 * \dfrac{Q_{Ker}(U,v)}{(J_{KerI}(U,v) - Q_{Ker}(U,v))/\gamma}$; For pKMFCM S-II,

   update $\gamma = 0.1 * \dfrac{Q_{Ker}(U,v)}{(J_{KerII}(U,v) - Q_{Ker}(U,v))/\gamma}$

3) Apply pKMFCM respectively to the $\Omega_c$.

4) Calculate $E_{Kerv}$ on validation set $\Omega_v$.

5) For pKMFCM S-I, set $\gamma = \gamma + 0.1 * \dfrac{Q_{Ker}(U,v)}{(J_{KerI}(U,v) - Q_{Ker}(U,v))/\gamma}$; For pKMFCM S-II,

   set $\gamma = \gamma + 0.1 * \dfrac{Q_{Ker}(U,v)}{(J_{KerII}(U,v) - Q_{Ker}(U,v))/\gamma}$. Go to step 3).

6) Select the $\gamma = \arg\min_\gamma \{E_{Kerv}\}$ within maximum updating times $T_\gamma$.

7) Apply pKMFCM with the optimized $\gamma$ to the entire data set $\Omega$.

The update processing of KMFCM algorithms is $\gamma = 0$ in the **Algorithm 1 & 2** of pKMFCM, while pMFCM algorithm can be treated as linear kernel (also a special case of Poly kernel) in KMFCM. MFCM algorithm is pMFCM with penalty factor $\gamma = 0$. To accelerate the computation speed, the initialization centroid $v_k^{(1)}$ of pKMFCM can be substituted by the corresponding $v_k^{(1)}$ of MFCM.

# 5. Experiments and Results

## 5.1. Database

To evaluate the performances of MFCM, pMFCM, KMFCM, pKMFCM algorithms, ten data sets from UCI Machine Learning and 3 images are involved in the experiments.

UCI Machine Learning data set is a standard benchmark database widely used in evaluation of pattern recognition and machine learning. It can be downloaded from the UCI repository of machine learning databases *www.ics.uci.edu/mlearn/MLrepository*. Ten data sets [9] involved in our experiments include: iris (I), wine (W), ionosphere (S), Wisconsin breast cancer (B), Wisconsin diagnostic breast cancer (WDBC), sonar mines versus rocks (O), glass (G), Haberman (H), ecoli protein localization sites (E), Pima Indians diabetes (D).

The first image file is a $64 \times 64$ pixels synthetic test image which is similar to the one used in [10,11,13], which posses two classes with intensity values of 0 and 128. The synthetic image is noised by 5% and 10% Gaussian noises and "salt andpepper" noises respectively. The images are shown in Fig. 2.

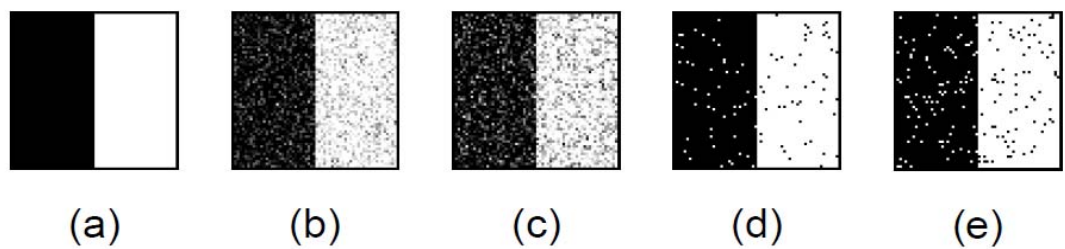

**Figure 2. The synthetic image 1 with noises. (a) original image. (b) 5% Gaussian noised image. (c) 10% Gaussian noised image. (d) 5% salt and pepper noised image. (e) 10% salt and pepper noised image.**

The second image file is a T1–weighted magnetic resonance brain image (MRI) file with slice thickness of 1mm, 3 % noise and no intensity inhomogeneities, which is downloaded from *http://mouldy.bic.mni.mcgill.ca/brainweb/* and is evaluated in [10,11,13]. We scale the original data to $0 \sim 255$ integers, and the original MRI. The image 3 are from "The Berkeley Segmentation Dataset and Benchmark", available at http://www.eecs.berkeley.edu/ Research/Projects/CS/vision/bsds/. Image 3 are converted to gray images. All of the original image data and corresponding gaussian kernel fitting [14] are shown in Fig. 3, where the peaks of kernel fitting are utilized to determine the centroids of images for fuzzy clustering.

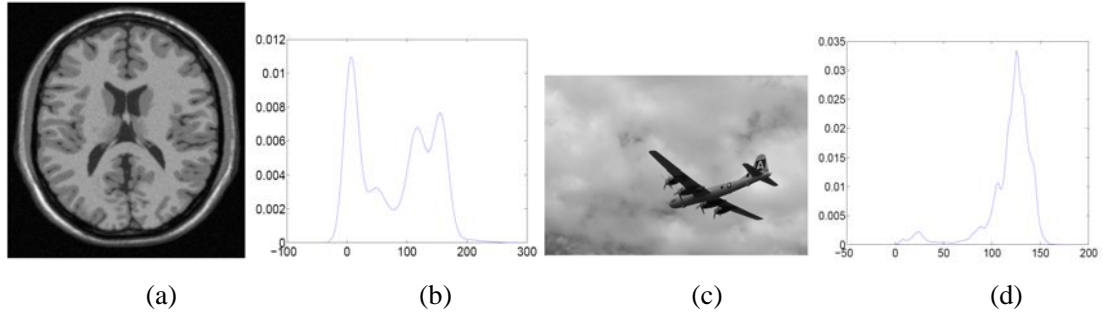

|  (a)  |  (b)  |  (c)  |  (d)  |

**Figure 3. (a) The synthetic T1-weighted MRI. (b) Kernel fitting of MRI. (c)Berkeley plane (d) Kernel fitting of Berkeley plane.**

*5.2. Classification Criterion*

The traditional classification criterion is based on the maximization of membership function, that is the sample $x_n$ is classified into the $k$-th centroid, if

$$k = \arg \max_{1 \leq j \leq C} u_{jn}. \qquad (42)$$

Also, segmentation accuracy [10], a criterion index, is employed to evaluate each of MFCM, pMFCM, KMFCM and pKMFCM algorithms, which is

$$\text{SA} = \frac{\text{number of pixels correctly classified}}{\text{total number of pixels}} \times 100\% \qquad (43)$$

For the pattern recognition of UCI dataset, the pixels in the definition of SA for pixels should be revised to the samples of the time series data.

*5.3. Experimental Results*

**5.3.1. Model parameters of MFCM, pMFCM and pKMFCM**

In our framework of robust statistic based MFCM, pMFCM, KMFCM and pKMFCM algorithms, 3 kernel functions (Poly, RBF ,Tanh), 7 weight functions , two types of spatial constrained penalty choices for pMFCM algorithm (S-I and S-II), and two choices of neighborhood information for image (Fig.1(b)(c)) ( nn-I and nn-II) are involved in experiments. For time series data, the neighborhood information will be the same type shown in Fig.1(a). There are totally 20 pKMFCM models except the fuzzy index $m$.

To evaluate the model selections of weight functions, penalty functions, neighborhood information and fuzziness indexes, various possible combinations make the evaluation process a burden of computation, the experiments outline is designed as follows:

For UCI datasets, in all the experiments, the parameters are set as follows: the maximum updating times $T = 20$, $T_\gamma = 10$, and $\varepsilon = 10^{-5}$. Given data set $\{x_1,\cdots,x_N\}$, set the diameter of data $\sigma = \max_{1\leq i \leq N} \max_{1 \leq j \leq N} \| x_i - x_j \|$. In the evaluation of MFCM and pMFCM model, we set $m \in \{1.2, 1.4, 1.6, \cdots, 3\}$, $\beta \in \{1,2,3\} * \sigma$. For the poly kernel KMFCM and pKMFCM, $(\beta, \theta, d) = \{1,2,3\} * \sigma^2 \times \{0.1, 1\} \times \{2, 4\}$. For the RBF kernel KMFCM and pKMFCM, $\beta = \{1, 2\} * \sigma^2$. For the tanh kernel KMFCM and pKMFCM, $(\beta, \theta) = \{1, 2\} * \sigma^2 \times \{0.1\}$. Each model is update 20 times, and the best result is reported in pattern recognition tasks.

To demonstrate the image segmentation effectiveness, image 1 –image 3 are investigated for image segmentation tasks with pMFCM and pKMFCM algorithms.

All the experiments are implemented with Matlab (MathWorks,Natick,MA) in a Windows XP system (Microsoft, Redmond, WA). All the experiments are run on a Dell® Optiplex 780 computer with Intel(R) Core(TM)2 Quad CPU Q9400 @2.66GHz and 4GB RAM.

### 5.3.2. Classification performance of MFCM, pMFCM and pKMFCM on UCI dataset

To verify the pre–processing effect on fuzzy clustering, three pre–processing methods are investigated in the experiments:
1) Each dimension of data are normalized to normal distribution with zero mean and unit standard deviation, denoted as "N01".
2) Original data without pre–processing, denoted as "NoP".
3) Each dimension of data are scaled to [0,1] interval, denoted as "U01".

Conveniently, we adopt a notation for model description (model name, penalty, kernel, neighbor,pre-processing, weight). Each model of MFCM, pMFCM, KMFCM and pKMFCM is performed 20 times, the same random initialization of $U^{(0)}$ is applied to each model, and the best classification accuracy rates are reported. We set $\Omega_c = \Omega_v = \Omega$ in optimizing $\gamma$ in pMFCM and pKMFCM.

To illustrate the effects of pre–processing methods and fuzzy index $m$, the average performances of MFCM, pMFCM, KMFCM and pKMFCM with $L_2$ weight function on 10 UCI datasets along fuzzy index are listed in Figure 4. While the investigated model is with multiple parameters, the best results are reported among all cases.

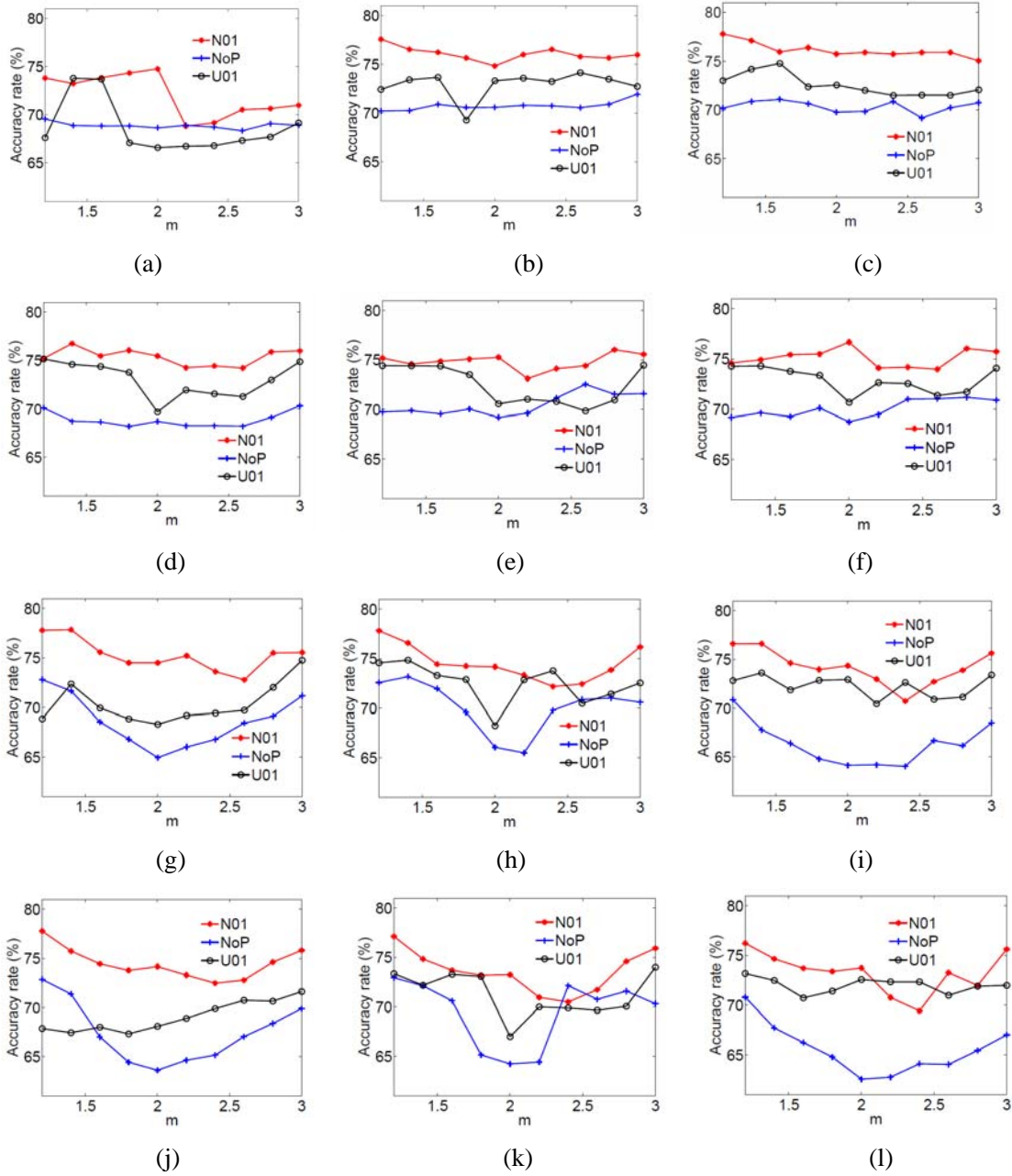

**Figure 4.** The average accuracy rates of MFCM, pMFCM, KMFCM, pKMFCM along $m$ with N01, NoP, U01 pre-processing on UCI datasets. (a) MFCM  (b) pMFCM S-I  (c) pMFCM S-II  (d) KMFCM poly  (e) KMFCM RBF  (f) KMFCM Tanh  (g) pKMFCM S-I poly  (h) pKMFCM S-I RBF  (i) pKMFCM S-I Tanh  (j) pKMFCM S-II poly  (k) pKMFCM S-II RBF (l) pKMFCM S-II Tanh.

As MFCM, pMFCM, KMFCM and pKMFCM with $L_2$ weight function are equivalent to the traditional FCM, SFCM, KFCM and SKFCM, Figure 4 shows that N01 pre-processing has the best performance on MFCM, pMFCM, KMFCM and pKMFCM with $L_2$ weight function on most fuzzy indexes. And U01 pre-processing performs better than NoP on some of the fuzzy indexes. To compare the effects of all 7 weight functions on MFCM, pMFCM, KMFCM and pKMFCM, the average of best results on 10 UCI datasets within all the fuzzy indexes are listed on Table 1 (The detail model parameters are in Appendix).

**Table 1.** The classification accuracy (%) of MFCM, pMFCM, KMFCM, pKMFCM algorithms on UCI datasets.

| Model | Pre-Pro | $L_2$ | $L_1-L_2$ | Huber | GM | Welsch | Cauchy | Fair |
|---|---|---|---|---|---|---|---|---|
| MFCM | N01 | 74.77 | 75.41 | 74.82 | 58.84 | **75.45** | 74.79 | 74.84 |
|  | NoP | 69.56 | 69.79 | 69.60 | 57.54 | **69.69** | **69.69** | 69.56 |
|  | U01 | 73.80 | 74.48 | 74.27 | 73.20 | **74.33** | **74.33** | 74.31 |
| pMFCM S-I | N01 | 77.59 | 77.31 | 78.91 | 59.22 | 80.19 | 80.23 | **80.45** |
|  | NoP | 71.94 | 73.24 | 72.20 | 62.00 | 73.55 | 73.21 | **74.70** |
|  | U01 | 74.14 | 75.12 | **75.24** | 72.72 | 74.73 | 74.83 | 75.08 |
| pMFCM S-II | N01 | 77.80 | 77.22 | 78.14 | 58.89 | 79.09 | **79.44** | 79.20 |
|  | NoP | 71.06 | 72.38 | 71.95 | 59.31 | 72.49 | 72.55 | **72.95** |
|  | U01 | 74.77 | 74.03 | 73.51 | 71.47 | 74.09 | 74.20 | **75.04** |
| KMFCM Poly | N01 | 76.75 | 77.37 | 77.87 | **78.26** | 77.46 | 77.31 | 78.03 |
|  | NoP | 70.33 | **71.77** | 69.70 | 71.35 | 70.10 | 70.10 | 70.60 |
|  | U01 | 75.17 | 74.93 | 75.03 | 75.15 | 74.86 | 75.14 | **75.79** |
| KMFCM RBF | N01 | 76.03 | 76.49 | **78.03** | 75.85 | 77.01 | 76.78 | 77.41 |
|  | NoP | 72.52 | 71.07 | 71.53 | 72.10 | 70.11 | 70.11 | 69.76 |
|  | U01 | 74.47 | 74.34 | **75.56** | 74.83 | 74.22 | 74.52 | 74.67 |
| KMFCM Tanh | N01 | 76.66 | 76.25 | **78.35** | 75.98 | 77.16 | 77.26 | 77.39 |
|  | NoP | 71.17 | 72.26 | 71.24 | **72.47** | 69.81 | 69.97 | 70.72 |
|  | U01 | 74.29 | 75.55 | 75.17 | 74.92 | 74.98 | 75.11 | **75.64** |
| pKMFCM S-I Poly | N01 | 77.85 | 76.08 | 78.53 | 70.86 | 78.65 | 78.73 | **79.57** |
|  | NoP | 72.80 | 70.36 | 72.56 | 65.69 | 72.34 | **73.04** | 72.84 |
|  | U01 | 74.76 | 76.22 | 76.03 | 75.91 | 74.40 | 74.34 | **76.31** |
| pKMFCM S-I RBF | N01 | 77.83 | 75.31 | **78.70** | 69.55 | 78.28 | 77.99 | 78.22 |
|  | NoP | 73.17 | 71.74 | 73.02 | 67.65 | 72.85 | 72.24 | 72.97 |
|  | U01 | 74.84 | 76.67 | **77.95** | 71.57 | 77.26 | 76.45 | 76.54 |
| pKMFCM S-I Tanh | N01 | 76.61 | 74.61 | 77.37 | 69.77 | 78.09 | 77.97 | **78.20** |
|  | NoP | 70.88 | 66.29 | 69.36 | 65.66 | **72.14** | 69.46 | 70.37 |
|  | U01 | 73.63 | 75.08 | **76.15** | 70.83 | 75.56 | 75.41 | 75.55 |
| pKMFCM S-II Poly | N01 | 77.78 | 75.49 | 78.22 | 71.43 | 78.36 | **78.42** | 78.32 |
|  | NoP | 72.86 | 69.69 | **73.10** | 63.75 | 72.59 | 72.56 | 72.58 |
|  | U01 | 71.63 | **75.34** | 71.78 | 72.29 | 73.08 | 72.62 | 74.28 |
| pKMFCM S-II RBF | N01 | 77.12 | 74.23 | 76.77 | 67.46 | 77.44 | 77.18 | **77.46** |
|  | NoP | 72.94 | 70.57 | 72.60 | 66.71 | **73.52** | 72.77 | 72.60 |
|  | U01 | 74.00 | 74.66 | 75.51 | 71.20 | 75.47 | **76.65** | 74.62 |
| pKMFCM S-II Tanh | N01 | 76.23 | 72.86 | 77.06 | 67.60 | **77.60** | 77.20 | 77.34 |
|  | NoP | 70.79 | 65.86 | 69.32 | 61.15 | **72.75** | 69.43 | 71.18 |
|  | U01 | 73.17 | 74.61 | 74.40 | 72.74 | 74.44 | **75.30** | 74.47 |
| Average |  | 74.21 | 73.74 | 74.71 | 69.05 | 74.84 | 74.65 | 74.99 |

From Table 1, the conclusion can be reached that,
- Except (KMFCM,RBF,NoP) and (pKMFCM,S-I,RBF,NoP), there exist at least one weight model who has better performance than baseline with weight $L_2$. This conclusion demonstrates the effectiveness of our extension of robust statistic based MFCM from FCM. The weak performances of above two models also appear in the Table 8 of [9], where KFCM-F(G) has slightly worse performance than FCM on W, C and E on UCI datasets, where they perform with N01 pre-processing method.
- Under the addressing statistical criterion, the S-I method is slightly better than S-II.
- Among three pre-processing methods, U01 is slightly better than NoP and U01.
- Taking the three kernels into account, comparing the average rates of KMFCM and pKMFCM, the proper order would be
    - $(KMFCM, Poly) \succ (KMFCM, Tanh) \succ (KMFCM, RBF)$;
    - $(pKMFCM, S-I, Poly) \succ (pKMFCM, S-I, Tanh) \succ (pKMFCM, S-I, RBF)$;
    - $(pKMFCM, S-II, Poly) \succ (pKMFCM, S-II, Tanh) \succ (pKMFCM, S-II, RBF)$.

    However, there is more parameter choice in Poly kernel based models than other two kernels. Poly kernel is better than RBF kernel is also founded in Table 8 in [9] on UCI data set. It is true on our M-estimation based fuzzy clustering models.
- If the average accuracy rate on all models with same weight function is considered, an appropriate choice would be, from the best to the worst,

    $$Fair \succ Welsch \succ Huber \succ Cauchy \succ L_2 \succ L_1 - L_2 \succ GM$$

    This is a statistical conclusion, when a specific data set is considered, the best model for it could lead poor performance to the other data set, because of the difference of data property in sample space.

The above discussion is based on the average on all the 10 UCI datasets on same parameters of a given model. If all the models are taken into account, we give the baseline results with N01 pre-processing and KFCM with RBF under N01, which are denoted as (MFCM,N01,$L_2$) and (KMFCM,RBF,N01,$L_2$) respectively, and the best results among all the candidate models are listed in Table 2. As this discussion avoids the case that for a given parameter of a fixed model, the performance of the model with the given parameter achieves better performance on one data set, while worse performance on the others. The expected upper bound will reveal the best recognition on the tested data sets among all our extended fuzzy clustering models and pre-processing methods.

According to the average rates from M1 to M12 in Table 2, pKMFCM model can achieve higher performance than KMFCM, pMFCM and MFCM at the cost of so many parameter combinations. If all the models, parameters and pre-processing methods are involved in candidate of model optimization neglecting the specific model framework, the average of maximum recognition rateson 10 UCI data sets can reach 90.07%. Comparing the best average results in Table 2 with Table 1, there is large gap in recognition rate values. This phenomenon manifests that model optimization still a challenge problem, especially in the cases that our pre-processing methods, weight functions, kernel mapping functions, and penalty functions are nonlinear. The upper bound expectant recognition rate points out the optimization directory of pattern recognition .

As for the computation speed, the CPU time depends on the complexity of optimization design, taking Iris (with N01) for example, for one parameter of each MFCM, pMFCM, KMFCM and pKMFCM, the average CPU times (second) ranges are M1: $0.19 \sim 0.29$ s; M2: $0.39 \sim 0.57$ s; M3: $0.51 \sim 0.70$ s; M4: $0.09 \sim 0.35$ s; M5: $0.14 \sim 0.17$ s; M6: $0.23 \sim 0.39$ s; M7: $1.03 \sim 1.33$ s; M8: $0.44 \sim 0.73$ s; M9: $1.19 \sim 1.72$ s; M10: $1.26 \sim 2.00$ s; M11: $0.61 \sim 0.89$ s; M12: $1.48 \sim 1.81$ s. The computation CPU time of one test data finally depends on data size, model combinations, model parameters, stop criteria, and optimization strategy, which lead to the computation complexity for optimization in pattern recognition.

**Table 2.** The upper bound classification accuracy (%) of MFCM, pMFCM, KMFCM, pKMFCM algorithms on 10 UCI datasets.

|  | I | W | S | B | WDBC | O | G | H | E | D | Average |
|---|---|---|---|---|---|---|---|---|---|---|---|
| M1 [a] | 92.67 | 97.19 | 71.79 | 96.57 | 93.85 | 65.87 | 60.75 | 59.15 | 73.81 | 72.27 | 78.39 |
| M2 | 100.00 | 100.00 | 69.52 | 96.71 | 93.15 | 89.90 | 73.83 | 62.75 | 81.55 | 70.44 | 83.79 |
| M3 | 100.00 | 100.00 | 69.52 | 96.85 | 93.15 | 86.06 | 71.50 | 62.42 | 81.25 | 70.31 | 83.11 |
| M4 | 94.67 | 98.31 | 78.63 | 97.42 | 95.96 | 74.52 | 72.90 | 75.82 | 77.08 | 74.61 | 83.99 |
| M5 | 97.33 | 97.19 | 78.06 | 97.42 | 95.25 | 74.04 | 71.03 | 73.53 | 75.00 | 73.70 | 83.26 |
| M6 | 98.00 | 97.19 | 78.35 | 97.57 | 95.43 | 73.08 | 71.03 | 75.49 | 72.32 | 74.09 | 83.26 |
| M7 | 100.00 | 100.00 | 95.16 | 95.85 | 94.73 | 82.69 | 84.58 | 74.84 | 83.93 | 73.05 | 88.48 |
| M8 | 100.00 | 100.00 | 82.91 | 89.27 | 94.38 | 82.21 | 85.05 | 68.30 | 80.95 | 71.09 | 85.42 |
| M9 | 100.00 | 100.00 | 90.88 | 92.13 | 94.02 | 81.73 | 85.05 | 73.53 | 86.31 | 71.09 | 87.47 |
| M10 | 100.00 | 100.00 | 95.44 | 95.71 | 95.25 | 82.69 | 80.84 | 74.51 | 82.74 | 71.22 | 87.84 |
| M11 | 100.00 | 100.00 | 84.62 | 89.13 | 94.20 | 82.21 | 82.24 | 68.95 | 81.85 | 71.09 | 85.43 |
| M12 | 100.00 | 100.00 | 92.02 | 89.99 | 94.02 | 81.73 | 73.36 | 73.53 | 85.71 | 71.09 | 86.15 |
| Maximum | 100.00 | 100.00 | 95.44 | 97.57 | 95.96 | 89.90 | 85.05 | 75.82 | 86.31 | 74.61 | 90.07 |

### 5.3.3. Image segmentation of synthetic image with MFCM, pMFCM, KMFCM and pKMFCM

In this experiment, the performances of MFCM, pMFCM, KMFCM and pKMFCM algorithms are evaluated on the artificial image 1 and the noise destroyed images with 5% and 10% "Gaussian noise" and "salt and pepper noise" respectively. To compare the results in [11,13], we adopt $\gamma = 3.8$ and $m = 2$ in all of the penalty models of pMFCM and pKMFCM algorithms with S-I or S-II penalty, and nn-I or nn-II neighbor information. For nn-I neighbor information, $N_R = 4$; and for nn-II neighbor information, $N_R = 8$.

To demonstrate the segmentation effectiveness, we first evaluate the image segmentation on noised images and then filter the noised image with $3 \times 3$ window mean filter and $3 \times 3$

---

[a] Model abbreviation: M1: MFCM; M2: pMFCM S-I; M3: pMFCM S-II; M4: KMFCM Poly; M5: KMFCM RBF; M6: KMFCM Tanh; M7: pKMFCM S-I Poly; M8: pKMFCM S-I RBF; M9: pKMFCM S-I Tanh; M10: pKMFCM S-II Poly; M11: pKMFCM S-II RBF; M12: pKMFCM S-II Tanh

median filter respectively as [13].

Also, each model is performed 10 times , the same initialization of $U^{(0)}$ is applied to every model, and the best classification accuracy results are reported. The experimental results with 5% and 10% of two kinds of noises are list in Table 3 – Table 6 (Footnote under the same items means having the same results.).

**Table 3.** The *SA*(%) of image 1 (5% Gaussian noises) with MFCM,KMFCM, pMFCM and pKMFCM.

| Model | neighbor | $L_2$ | $L_1 - L_2$ | Huber | GM | Welsch | Cauchy | Fair |
|---|---|---|---|---|---|---|---|---|
| MFCM | | 98.54 | 98.54 | 98.54 | 98.54 | 98.54 | 98.54 | 98.54 |
| pMFCM | nn-I | 94.75 | 99.93 | 99.93 | 98.88 | 99.93 | 99.93 | 99.93 |
| S-I [a] | nn-II | 99.95 | 99.95 | 99.95 | 99.95 | 99.95 | 99.95 | 99.95 |
| KMFCM Poly | | 98.56 | 98.54 | 98.56 | 98.54 | 98.56 | 98.56 | 98.56 |
| KMFCM RBF | | 98.54 | 98.54 | 98.54 | 98.56 | 98.54 | 98.54 | 98.54 |
| KMFCM Tanh | | 98.56 | 98.56 | 98.56 | 98.56 | 98.56 | 98.56 | 98.56 |
| pKMFCM S-I Poly [b] | nn-I | 99.61 | 99.39 | 99.39 | 99.46 | 99.39 | 99.39 | 99.39 |
| pKMFCM S-I RBF [c] | nn-I | 99.58 | 99.39 | 99.39 | 99.46 | 99.39 | 99.39 | 99.39 |
| pKMFCM | nn-I | 99.78 | 99.68 | 99.58 | 99.68 | 99.63 | 99.39 | 99.39 |
| S-II RBF | nn-II | 99.90 | 99.76 | 99.71 | 99.80 | 99.76 | 99.39 | 99.39 |

**Table 4.** The *SA*(%)of image 1 (10% Gaussian noises) with MFCM,KMFCM, pMFCM and pKMFCM.

| Model | neighbor | $L_2$ | $L_1 - L_2$ | Huber | GM | Welsch | Cauchy | Fair |
|---|---|---|---|---|---|---|---|---|
| MFCM | | 94.24 | 94.24 | 94.24 | 94.24 | 94.24 | 94.24 | 94.24 |
| pMFCM | nn-I | 85.64 | 99.56 | 99.56 | 94.38 | 99.56 | 99.56 | 99.56 |
| S-I [d] | nn-II | 99.61 | 99.56 | 99.51 | 99.63 | 99.51 | 99.51 | 99.51 |
| KMFCM Poly [e] | | 94.24 | 94.24 | 94.24 | 94.24 | 94.24 | 94.24 | 94.24 |
| pKMFCM | nn-I | 97.61 | 96.97 | 96.92 | 97.22 | 96.92 | 96.92 | 96.95 |
| S-I Poly [f] | nn-II | 97.61 | 96.97 | 96.92 | 97.31 | 96.92 | 96.92 | 96.95 |
| pKMFCM S-I RBF [g] | nn-I | 97.61 | 96.97 | 96.92 | 97.17 | 96.92 | 96.92 | 96.95 |
| pKMFCM S-II RBF [h] | nn-I | 97.61 | 96.97 | 96.92 | 97.19 | 96.92 | 96.92 | 96.95 |

---

[a] (pMFCM,S-II)
[b] (pKMFCM,S-I,Poly,nn-II)
[c] (pKMFCM,S-I,RBF,nn-II),(pKMFCM,S-I,Tanh),(pKMFCM,S-II,Poly),(pKMFCM,S-II,Tanh)
[d] (pMFCM,S-II)
[e] (KMFCM,RBF), (KMFCM,Tanh).
[f] (pKMFCM,S-I,Tanh),(pKMFCM,S-II,Poly),(pKMFCM,S-II,Tanh)
[g] (pKMFCM,S-I,RBF,nn-II)
[h] (pKMFCM,S-II,RBF,nn-II)

**Table 5.** The *SA*(%) of image 1 (5% salt and pepper noises) with MFCM,KMFCM, pMFCM and pKMFCM.

| Model[a] | neighbor | $L_2$ | $L_1 - L_2$ | Huber | GM | Welsch | Cauchy | Fair |
|---|---|---|---|---|---|---|---|---|
| MFCM | | 97.53 | 97.53 | 97.53 | 97.53 | 97.53 | 97.53 | 97.53 |
| pMFCM S-I | nn-I | 99.58 | 99.56 | 99.56 | 99.63 | 99.56 | 99.56 | 99.56 |
| | nn-II | 97.53 | 97.53 | 97.53 | 97.53 | 97.53 | 97.53 | 97.53 |
| pMFCM S-II[b] | nn-I | 99.58 | 99.56 | 99.56 | 99.63 | 99.56 | 99.56 | 99.56 |
| | nn-II | 97.53 | 97.53 | 97.53 | 97.56 | 97.53 | 97.53 | 97.53 |
| pKMFCM S-I Poly[c] | nn-I | 99.44 | 99.44 | 99.44 | 99.46 | 99.44 | 99.44 | 99.44 |

**Table 6.** The *SA*(%) of image 1 (10% salt and pepper noises) with MFCM,KMFCM, pMFCM and pKMFCM.

| Model | neighbor | $L_2$ | $L_1 - L_2$ | Huber | GM | Welsch | Cauchy | Fair |
|---|---|---|---|---|---|---|---|---|
| MFCM[d] | | 95.65 | 95.65 | 95.65 | 95.65 | 95.65 | 95.65 | 95.65 |
| pMFCM S-I[e] | nn-I | 99.05 | 98.93 | 98.90 | 99.12 | 98.90 | 98.90 | 98.90 |
| | nn-II | 95.65 | 95.65 | 95.65 | 98.85 | 95.65 | 95.65 | 95.65 |
| pKMFCM S-I Poly[f] | nn-I | 98.71 | 98.71 | 98.68 | 98.78 | 98.68 | 98.68 | 98.68 |
| | nn-II | 98.71 | 98.71 | 98.68 | 98.75 | 98.68 | 98.68 | 98.68 |

The performances on image 1 with 5% and 10% Gaussian noises show that pMFCM achieve the best *SA*, and the pKMFCM is more accurate than KMFCM and MFCM ( KMFCM and MFCM have same performances.). Except $L_2$ case, the different between nn-I and nn-II is slight within pMFCM models. For pKMFCM, nn-I and nn-II have same effects to recognition rate. For the image 1 with 5% and 10% "salt and pepper" noises, MFCM,KMFCM and (pKMFCM,nn-II) have almost same performance. Fathermore, all pKMFCM models have the better SA values, and (pKMFCM,nn-I) reach the best SA value.

As mean filter and median filter are appropriate tools for smoothing Gaussian noise and "salt and pepper" noise respectively [11, 13], we adopt a $3 \times 3$ window around the considered pixel for both mean filter and median filter, and the segmentationtasks are evaluated by MFCM, pMFCM, KMFCM, and pKMFCM algorithms too. The segmentation accuracies are listed from Table 7 to Table 10.

---

[a] Footnote under the same items means having the same results.
[b] (KMFCM,Poly),(KMFCM,RBF),(KMFCM,Tanh)
[c] (pKMFCM,S-I,Poly,nn-II),(pKMFCM,S-I,RBF), (pKMFCM,S-I,Tanh), (pKMFCM,S-II,Poly), (pKMFCM,S-II,RBF), (pKMFCM,S-II,Tanh).
[d] (KMFCM,Poly), (KMFCM,RBF), (KMFCM,Tanh)
[e] (pMFCM,S-II)
[f] (pKMFCM,S-I,RBF), (pKMFCM,S-I,Tanh), (pKMFCM,S-II,Poly), (pKMFCM,S-II,RBF), (pKMFCM,S-II, Tanh).

**Table 7.** The $SA(\%)$ of image 1 (5% Gaussian noise) with MFCM, KMFCM, pMFCM and pKMFCM.

| Model | neighbor | $L_2$ | $L_1-L_2$ | Huber | GM | Welsch | Cauchy | |
|---|---|---|---|---|---|---|---|---|
| MFCM [a] | | 99.90 | 99.90 | 99.90 | 99.90 | 99.90 | 99.90 | 99.90 |
| pMFCM | nn-I | 91.41 | 99.80 | 99.80 | 98.46 | 99.80 | 99.80 | 99.80 |
| S-I [b] | nn-II | 99.90 | 99.93 | 99.93 | 99.93 | 99.93 | 99.93 | 99.93 |
| pKMFCM S-I Poly [c] | nn-I | 99.93 | 99.93 | 99.93 | 99.93 | 99.93 | 99.93 | 99.93 |

**Table 8.** The $SA(\%)$ of image 1 (10% Gaussian noise) with MFCM, KMFCM, pMFCM and pKMFCM.

| Model | neighbor | $L_2$ | $L_1-L_2$ | Huber | GM | Welsch | Cauchy | Fair |
|---|---|---|---|---|---|---|---|---|
| MFCM | | 99.46 | 99.46 | 99.46 | 99.46 | 99.46 | 99.46 | 99.46 |
| pMFCM | nn-I | 84.89 | 99.12 | 99.17 | 94.87 | 99.17 | 99.17 | 99.17 |
| S-I [d] | nn-II | 99.46 | 99.56 | 99.56 | 99.61 | 99.56 | 99.56 | 99.56 |
| KMFCM Poly | | 99.54 | 99.46 | 99.54 | 99.54 | 99.54 | 99.54 | 99.51 |
| KMFCM RBF | | 99.54 | 99.54 | 99.51 | 99.54 | 99.51 | 99.51 | 99.51 |
| KMFCM Tanh | | 99.54 | 99.49 | 99.54 | 99.54 | 99.54 | 99.54 | 99.46 |
| pKMFCM S-I Poly [e] | nn-I | 99.63 | 99.58 | 99.56 | 99.58 | 99.56 | 99.56 | 99.56 |

**Table 9.** The $SA(\%)$ of image 1 (5% salt and pepper noise) with MFCM, KMFCM, pMFCM and pKMFCM.

| Model | neighbor | $L_2$ | $L_1-L_2$ | Huber | GM | Welsch | Cauchy | Fair |
|---|---|---|---|---|---|---|---|---|
| MFCM [f] | | 99.90 | 99.90 | 99.90 | 99.90 | 99.90 | 99.90 | 99.90 |
| pMFCM S-I [g] | nn-I | 99.90 | 99.90 | 99.90 | 99.93 | 99.90 | 99.90 | 99.90 |
| pKMFCM S-I Poly [h] | nn-I | 99.93 | 99.93 | 99.93 | 99.93 | 99.93 | 99.93 | 99.93 |

**Table 10.** The $SA(\%)$ of image 1 (10% salt and pepper noise) with MFCM, KMFCM, pMFCM and pKMFCM.

| Model | neighbor | $L_2$ | $L_1-L_2$ | Huber | GM | Welsch | Cauchy | Fair |
|---|---|---|---|---|---|---|---|---|
| MFCM [i] | | 99.78 | 99.78 | 99.78 | 99.78 | 99.78 | 99.78 | 99.78 |
| pKMFCM S-I Poly [j] | nn-I | 99.80 | 99.80 | 99.80 | 99.80 | 99.80 | 99.80 | 99.80 |

---

[a] (KMFCM,Poly),(KMFCM,RBF), (KMFCM,Tanh).
[b] (pMFCM,S-II)
[c] (pKMFCM,S-I,Poly,nn-II),(pKMFCM,S-I,RBF), (pKMFCM,S-I,Tanh), (pKMFCM,S-II,Poly), (pKMFCM,S-II,RBF), (pKMFCM,S-II,Tanh).
[d] (pMFCM S-II)
[e] (pKMFCM,S-I,Poly,nn-II), (pKMFCM,S-I,RBF), (pKMFCM,S-I,Tanh), (pKMFCM,S-II,Poly), (pKMFCM,S-II,RBF), (pKMFCM,S-II,Tanh).
[f] (KMFCM,Poly),(KMFCM,RBF), (KMFCM,Tanh) .
[g] (pMFCM,S-I,nn-II), (pMFCM,S-II)
[h] (pKMFCM,S-I,Poly,nn-II), (pKMFCM,S-I,RBF), (pKMFCM,S-I,Tanh), (pKMFCM,S-II,Poly), (pKMFCM,S-II,RBF), (pKMFCM,S-II,Tanh).
[i] (pMFCM,S-I), (pMFCM,S-II), (KMFCM,Poly), (KMFCM,RBF), (KMFCM,Tanh).
[j] (pKMFCM,S-I,Poly,nn-II), (pKMFCM,S-I,RBF), (pKMFCM,S-I,Tanh), (pKMFCM,S-II,Poly), (pKMFCM,S-II,RBF) and (pKMFCM,S-II,Tanh).

From Table 7 to Table 10, the pKMFCM models achieve best SA results among all MFCM, pMFCM,KMFCM and pKMFCM. This conclusion not only demonstrates the importance of smoothing filters, but also shows that pKMFCM has better performance that MFCM, pMFCM,KMFCM on the relatively "clean" images after filtering. In addition, nn-II has better performance than nn-I.

The experimental results from Table 1 to Table 10 show the performances of MFCM, pMFCM, KMFCM and pKMFCM, and they demonstrate the effectiveness of pKMFCM in pattern recognition. Generally, the introduction of weight function, penalty information, neighbor information and kernelization can improve the pattern recognition rate and segmentation accuracy.

### 5.3.4. Image segmentation of synthetic MRI with MFCM,KMFCM,pMFCM and pKMFCM

In this section, the image segmentation of image 2 (Fig.3(a)) are evaluated by MFCM, KMFCM, pMFCM and pKMFCM algorithms. The centroids number is set 3, and $\gamma = 0.1$ and $m = 2$. At first, we perform segmentation on original image , which means NoP, withall MFCM, KMFCM, pMFCM and pKMFCM. The experimental results show that GM weight function has poor performance even in MFCM model frame.Then, to find better strategy to overcome the deficiency, we exert pre-processing N01 on image 2, and redo the segmentation tasks with all the 20 pKMFCM models. The segmentation results show that all MFCM, pMFCM and KMFCM with possible weight functions perform well, however all pKMFCM models perform poorly. Finally, we scale the pixel values of image 2 to [0,4], and redo the experiments. The segmentation results are very similar to N01, the relatively successful segmentation results are shown in Fig.5.

We overcome the deficiency of (MFCM,GM) with N01 and Scaling pre-processing methods, yet the pKMFCM models thoroughly fail in segmentation with the above parameter setting. Due to scaling to [0,4], the different effects of weight functions are weaken, andthe discriminate among pixels is also reduced (Fig.3(b)), which leads to the failures of other models in MRI segmentation.

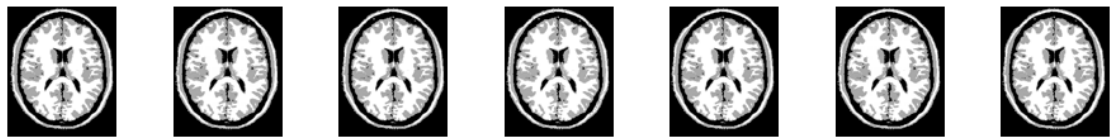

(a)

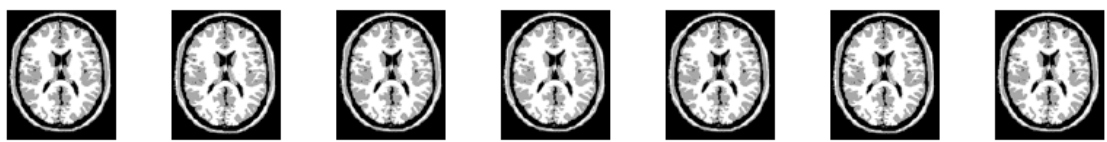

(b)

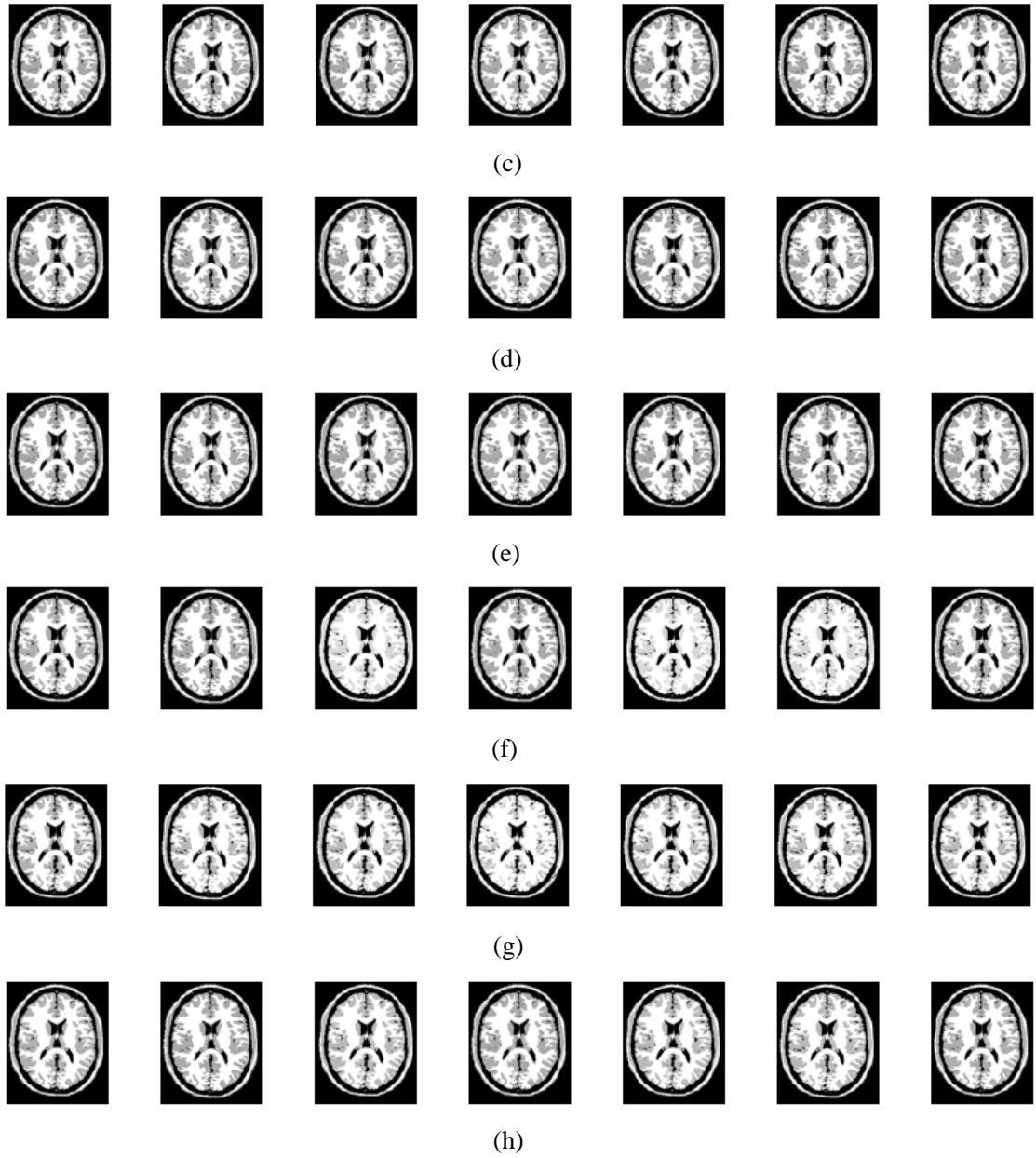

**Figure 5.** The image 2 MRI segmentation results. (a) MFCM (b) (pMFCM,S-I,nn-I) (c) (pMFCM,S-I,nn-II) (d) (pMFCM,S-II,nn-I) (e) (pMFCM,S-II,nn-II) (f) (KMFCM,Poly) (g) (KMFCM,RBF) (h) (KMFCM, Tanh). The subfigures from left to right are with different weightfunctions: $L_2$, $L_1-L_2$, Huber, GM, Welsch, Cauchy, Fair.

### 5.3.5. Image segmentation of Berkeley image with MFCM,KMFCM,pMFCM and pKMFCM

In this section, we segment image 3 (Fig.3(c)) with MFCM, pMFCM, KMFCM and pKMFCM without pre-processing. The centroid number is set 2 for all 20 MFCM, pMFCM, KMFCM and pKMFCM models. And, we set $\gamma = 0.1$, $m = 2$. Several cases, whatever failure or success, are shown in Fig 6.

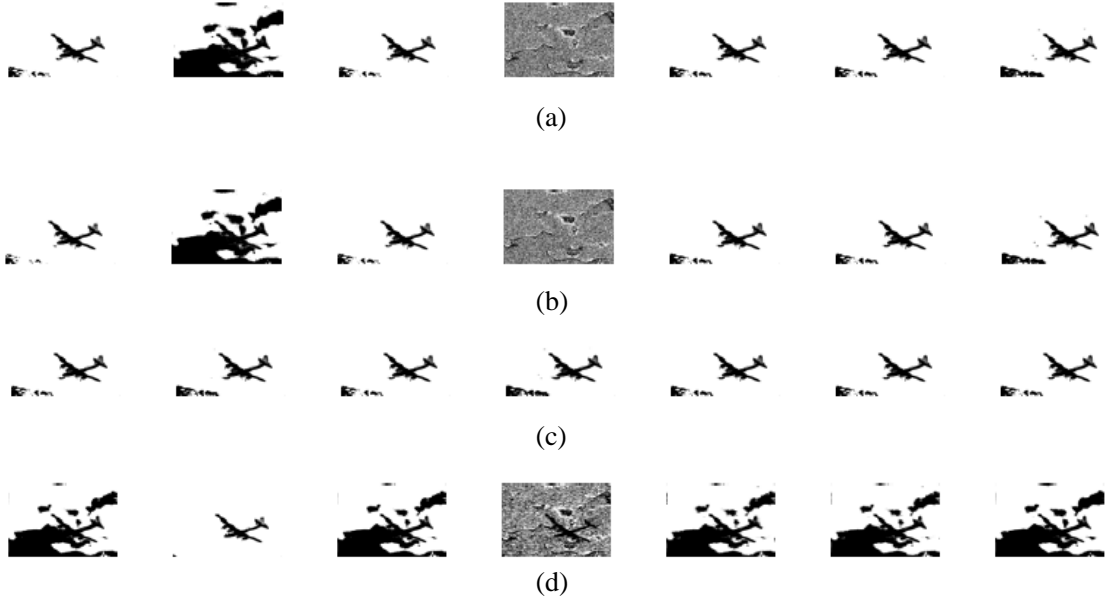

**Figure 6.** Segmentation results of image 3. (a) MFCM (b) (pMFCM,S-II,nn-II) (c) (KMFCM,RBF) (d) (pKMFCM, S-II,Tanh, nn-II). Each subfigure from left to right is with different weight functions: $L_2$, $L_1-L_2$, Huber, GM, Welsch, Cauchy, Fair.

As illustrated in Fig. 6, (KMFCM,RBF) has better performance than MFCM. Again, GM weight function takes bad performance in pMFCM and pKMFCM algorithms. And, (pKMFCM,Tanh,S-II,nn-II) with $L_1-L_2$ weight has achieved the best segmentation results, while the other weight functions perform badly. How to select the best model and parameter is still a difficult research work.

## 6. Conclusion

A framework of spatial penalty constrained MFCM algorithms is proposed and evaluated in pattern classification, noisy and real environment image segmentation on standard data sets. The effectiveness of weight function, penalty function, kernel function, neighborhood information and fuzziness index is also investigated. All of them take important roles in pattern recognition and image segmentation of pKMFCM. pKMFCM generalizes the traditional FCM and SFCM and SKFCM algorithms and provide many pMFCM and pKMFCM algorithms. Moreover, MFCM and KMFCM algorithms are the pMFCM and pKMFCM algorithms without penalty functions. The experimental results demonstrate the significance of our extension framework and show promising prospect in pattern recognition and image segmentation.

Within the discussed parameter ranges, pMFCM and pKMFCM have different performances in different tasks. The optimization of combination and parameter selection in the pMFCM and pKMFCM would be an important problems in model selections. Our experimental results are only limited to the investigated parameter ranges, while the parameter ranges are enlarged, the best experimental results could be expected more accurate, at the same time the computation consumption will rise. How to design the effective optimization among the extend fuzzy clustering models is an important problem to develop practical algorithm. We will discuss it in the future work, and extend the pMFCM to KFCM-K type extension. Finally, all of the pMFCM and pKMFCM algorithms will be applied to more pattern recognition, image segmentation and gene expression data analysis tasks.

## Acknowledgements

The authors would like to thank the reviewers for their valuable comments and suggestions, which make great improvement of this paper.

## References

[1] J. C. Bezdek. Pattern Recognition with Fuzzy Objective Function Algorithms. New York: Plenum Press. 1981.

[2] K.L. Wu, M.S. Yang. Alternative C-means Clustering Algorithm. Pattern Recognition. 35 (2002) 2267–2278.

[3] A. W. C. Liew, S. H. Leung, W. H. Lau. Fuzzy image clustering incorporating spatial continuity. IEE Proceedings Vision, Image and Signal Processing. 147(2) (2000) 185–192.

[4] D.,Dembele, P.,Kastner. Fuzzy C-means method for clustering microarray data. BIOINFORMATICS. 19(8) (2003) 973–980.

[5] R. N. Dave, R. Krishnapuram. Robust Clustering Methods: A Unified View. IEEE Trans. on Fuzzy Systems. 5(2) (1997) 270-293.

[6] Q. S. Cheng. Attribute Means Clustering. Systems Engineering-Theory & Practice. 9 (1998)124–126.

[7] J. Yu. General C–means Clustering Model. IEEE Transactions on pattern analysis and machine intelligence. 27(8) (2005) 1197–1211.

[8] J. W. Liu, M. Z. Xu. Bezdek Type Fuzzy Attribute C-means Clustering Algorithm. Journal of Beijing University of Aeronautics and Astronautics. 33 (2007) 1121–1126.

[9] D. Graves, W. Pedrycz. Kernel-based Fuzzy Clustering and Fuzzy Clustering: A comparative Experimental Study. Fuzzy Sets and Systems. 161(2010) 522–543.

[10] S. C. Chen, D. Q. Zhang. Robust Image Segmentation Using FCM With Spatial Constraints Based on New Kernel-Induced Distance Measure. IEEE Transactions on systems, man, and cybernetics –Part B: cybernetics. 34(4) (2004) 1907–1916.

[11] D.Q. Zhang, S.C. Chen. A Novel Kernelized Fuzzy C-Means Algorithm with Application in Medical Image Segmentation. Artificial Intelligence in Medicine, 32 (2004) 37–50.

[12] J. W. Liu, M. Z. Xu. Kernelized Fuzzy Attribute Means Clustering Algorithm. Fuzzy sets and systems. 159(18) (2008) 2428–2445.

[13] L. Chen, C.L.P. Chen., M.Z. Lu. A Multiple-Kernel Fuzzy C-Means Algorithm for Image

Segmentation. IEEE Transactions on Systems, Man, and Cybernetics—Part B: Cybernetics. 41(5) (2011) 1263 - 1274 .

[14] D. L. Pham, J. L. Prince. An adaptive fuzzy C–means algorithm for image segmentation in the presence of intensity inhomogeneities. Pattern Recognition Letters. 20 (1999) 57–68.

[15] D. L. Pham. Spatial Models for Fuzzy Clustering. Computer Vision and Image Understanding, 84 (2001) 285–297.

[16] M. N. Ahmed, S. M. Yamany, N. Mohamed, A. A. Farag, T. Moriarty. A modified fuzzy C-means algorithm for bias field estimation and segmentation of MRI data. IEEE Trans. Med. Imaging. 21 (2002) 193–199.

[17] L. Jiang, W. H. Yang. A Modified Fuzzy C-Means Algorithm for Segmentation of Magnetic Resonance Images. Proc. VIIth Digital Image Computing: Techniques and Applications, Sun C., Talbot H., Ourselin S. and Adriaansen T. (Eds.), Sydney, 10–12 Dec. 2003.

[18] K.S. Chuang, H.L. Tzeng, S. Chen, J. Wu, T.J. Chen. Fuzzy c-means clustering with spatial information for image segmentation. Computerized Medical Imaging and Graphics. 30 (2006) 9–15.

[19] D. Ozdemir, L. Akarun. A fuzzy algorithm for color quantization of images. Pattern Recognition. 35 (2002) 1785–1791.

[20] L. Zhu, F. L. Chung, S. T. Wang. Generalized fuzzy C–means clustering algorithm with improved fuzzy partitions. IEEE Transactions on Systems, Man, and Cybernetics– Part B: Cybernetics. 39(3) (2009) 578–591.

[21] P.J. Huber. Robust regression; asymptotics, conjectures and Monte Carlo, Ann. Stat., 1 (1973) 799–821.

[22] P.J. Huber. Robust Statistics, John Wiley, New York. 1981.

[23] Z.Y. Zhang. Parameter Estimation Techniques: A Tutorial with Application to Conic Fitting. Image and Vision Computing. 15(1) (1997) 59-76.

[24] L. Bhar. Robust Regression. In: Advances in Data Analytical Techniques. (eds.) R. Parsad, V.K. Gupta, L.M. Bhar, V.K. Bhatia. Indian Agricultural Statistics Research Institute. http://www.iasri.res.in /ebook/EBADAT /index.htm. Accessed 1 July 2010.

[25] Q.S. Cheng. Mathematical Principle of Digital Signal Processing (Second Edition). Beijing: Oil Industry Press. 1993. (in Chinese).

[26] J. Zhang. The Mean Field Theory in EM Procedures for Markov Random Fields. IEEE Transactions on Signal Processing. 40(10) (1992) 2570–2583.

[27] J. Shawe-Taylor, N. Cristianini. Kernel Methods for Pattern Analysis. Beijing: China Machine Press. 2005.

[28] T. Kanungo, D.M. Mount, N. Netanyahu, C. Piatko, R. Silverman, A.Y. Wu. An efficient k-means clustering algorithm: Analysis and implementation. Proc. IEEE Conf. Computer Vision and Pattern Recognition. (2002) 881–892.

[29] P. Arbelaez, M. Maire, C. Fowlkes, J. Malik. Contour Detection and Hierarchical Image

Segmentation. IEEE TPAMI, 33(5) (2011) 898-916.

[30] C.W. Hsu, C.C. Chang, C.J. Lin. A practical guide to support vector classification . Technical report, Department of Computer Science, National Taiwan University. July, 2003.

[31] B. Schölkopf, A.J. Smola. Learning with Kernels. MIT Press, Cambridge, MA, 2002.

[32] V. Vapnik. The Nature of Statistical Learning Theory. Springer Verlag, New York, 1995.

[33] T. Gartner. A Survey of Kernels for Structured Data. SIGKDD Exploration Newsletters. 5(1) (2003) 49–58.

[34] S. Zhou, J. Gan. Mercer kernel fuzzyc-means algorithm and prototypes of clusters. in:Proc. of Conf. on Internat. Data Engineering and Automated Learning,3177 (2004)613–618.

[35] P.R. Halmos. Naive Set Theory. Princeton, NJ: D. Van Nostrand Company, 1960.

[36] J.N. McDonald, N.A. Weiss. A Course in Real Analysis. Academic Press. 2004.

[37] M. Debruyne, A. Christmann, M. Hubert, J.A.K. Suykensd. Robustness of reweighted Least Squares Kernel Based Regression. Journal of Multivariate Analysis. 101 (2010) 447–463.

[38] H. Frigui, R. Krishnapuram. A Robust Algorithm for Automatic Extraction of an Unknown Number of Clusters from Noisy Data. Pattern Recognition Letters, 17(12)(1996) 1223–1232.

[39] C.H. Wang. Apply Robust Segmentation to the Service Industry Using Kernel Induced Fuzzy Clustering Techniques. Expert Systems with Applications. 37 (2010) 8395–8400.

[40] R. Winkler, F. Klawonn, R. Kruse. M-Estimator induced Fuzzy Clustering Algorithms. The 7th conference of the European Society for Fuzzy Logic and Technology (2011). 298–304.

# Appendix

Table A1. Detail parameters of MFCM, pMFCM,KMFCM,pKMFCM of best average rates on UCI datasets.

Table A1_1   weight   function   L2

|  | N01 | NoP | U01 |
|---|---|---|---|
| MFCM:: | 74.77 | 69.56 | 73.80 |
| m:: | 2.00 | 1.20 | 1.40 |
| pMFCM_S-I:: | 77.59 | 71.94 | 74.14 |
| m:: | 1.20 | 3.00 | 2.60 |
| pMFCM_S-II:: | 77.80 | 71.06 | 74.77 |
| m:: | 1.20 | 1.60 | 1.60 |
| KMFCM_Poly:: | 76.75 | 70.33 | 75.17 |
| m:: | 1.40 | 3.00 | 1.20 |
| Para: Poly $(\beta,\theta,d)$ | (3.0,0.1,4) | (3.0,1.0,2) | (3,0.1,2) |
| KMFCM_RBF:: | 76.03 | 72.52 | 74.47 |
| m:: | 2.80 | 2.60 | 3.00 |
| Para: RBF $\beta$ | 3.0 | 3.0 | 3.0 |
| KMFCM_Tanh:: | 76.66 | 71.17 | 74.29 |
| m:: | 2.00 | 2.80 | 1.40 |
| Para: Tanh $(\beta,\theta)$ | (3.0,1.0) | (3.0,1.0) | (3.0,1.0) |
| pKMFCM_S-I_Poly:: | 77.85 | 72.80 | 74.76 |
| m:: | 1.40 | 1.20 | 3.00 |
| Para: Poly $(\beta,\theta,d)$ | (1.0,1.0,4) | (3.0,1.0,4.0) | (1.0,1.0,2) |
| pKMFCM_S-I_RBF:: | 77.83 | 73.17 | 74.84 |
| m:: | 1.20 | 1.40 | 1.40 |
| Para: RBF $\beta$ | 1.0 | 1.0 | 1.0 |
| pKMFCM_S-I_Tanh:: | 76.61 | 70.88 | 73.62 |
| m:: | 1.40 | 1.20 | 1.40 |
| Para: Tanh $(\beta,\theta)$ | (1.0,0.1) | (2.0,0.1) | (2.0,0.1) |
| pKMFCM_S-II_Poly:: | 77.78 | 72.86 | 71.63 |
| m:: | 1.20 | 1.20 | 3.00 |
| Para: Poly $(\beta,\theta,d)$ | (1.0,1.0,2.0) | (3.0,1.0,4.0) | (1.0,1.0,2.0) |
| pKMFCM_S-II_RBF:: | 77.12 | 72.94 | 74.00 |
| m:: | 1.20 | 1.20 | 3.00 |
| Para: RBF $\beta$ | 1.0 | 3.0 | 1.0 |
| pKMFCM_S-II_Tanh:: | 76.23 | 70.79 | 73.17 |
| m:: | 1.20 | 1.20 | 1.20 |
| Para: Tanh $(\beta,\theta)$ | (1.0,0.1) | (3.0,0.1) | (1.0,0.1) |

Table A1_2  weight  function  L1-L2

|  | N01 | NoP | U01 |
|---|---|---|---|
| MFCM:: | 75.41 | 69.79 | 74.48 |
| m:: | 2.00 | 2.00 | 1.40 |
| pMFCM_S-I:: | 77.31 | 73.24 | 75.12 |
| m:: | 2.40 | 1.80 | 1.40 |
| pMFCM_S-II:: | 77.22 | 72.38 | 74.03 |
| m:: | 1.80 | 1.80 | 2.80 |
| KMFCM_Poly:: | 77.37 | 71.77 | 74.93 |
| m:: | 1.40 | 2.20 | 1.40 |
| Para: Poly $(\beta,\theta,d)$ | (3.0,0.1,4.0) | (3.0,0.1,2.0) | (3.0,1.0,2.0) |
| KMFCM_RBF:: | 76.49 | 71.07 | 74.34 |
| m:: | 2.00 | 3.00 | 1.80 |
| Para: RBF $\beta$ | 2.0 | 1.0 | 3.0 |
| KMFCM_Tanh:: | 76.25 | 72.26 | 75.55 |
| m:: | 2.00 | 1.80 | 1.40 |
| Para: Tanh $(\beta,\theta)$ | (2.0,0.1) | (3.0,1.0) | (3.0,1.0) |
| pKMFCM_S-I_Poly:: | 76.08 | 70.36 | 76.22 |
| m:: | 1.40 | 1.20 | 1.40 |
| Para: Poly $(\beta,\theta,d)$ | (1.0,1.0,2.0) | (1.0,1.0,2.0) | (2.0,1.0,2.0) |
| pKMFCM_S-I_RBF:: | 75.31 | 71.74 | 76.67 |
| m:: | 2.80 | 2.40 | 1.20 |
| Para: RBF $\beta$ | 1.0 | 1.0 | 1.0 |
| pKMFCM_S-I_Tanh:: | 74.61 | 66.29 | 75.08 |
| m:: | 1.80 | 2.20 | 3.00 |
| Para: Tanh $(\beta,\theta)$ | (1.0,0.1) | (2.0,0.1) | (1.0,0.1) |
| pKMFCM_S-II_Poly:: | 75.49 | 69.69 | 75.34 |
| m:: | 1.20 | 1.40 | 1.20 |
| Para: Poly $(\beta,\theta,d)$ | (1.0,1.0,4.0) | (1.0,1.0,2.0) | (1.0,1.0,4.0) |
| pKMFCM_S-II_RBF:: | 74.23 | 70.57 | 74.66 |
| m:: | 2.80 | 2.40 | 1.20 |
| Para: RBF $\beta$ | 1.0 | 1.0 | 1.0 |
| pKMFCM_S-II_Tanh:: | 72.86 | 65.86 | 74.61 |
| m:: | 3.00 | 2.20 | 1.40 |
| Para: Tanh $(\beta,\theta)$ | (1.0,0.1,1.0) | (2.0,1.0,1.0) | (2.0,1.0,1.0) |

Table A1_3  weight function   Huber

|  | N01 | NoP | U01 |
|---|---|---|---|
| MFCM:: | 74.82 | 69.60 | 74.27 |
| m:: | 2.00 | 1.40 | 1.40 |
| Para: Huber $\beta$ | 1 | 1 | 1 |
| pMFCM_S-I:: | 78.91 | 72.20 | 75.24 |
| m:: | 1.40 | 1.20 | 1.60 |
| Para: Huber $\beta$ | 1 | 1 | 1 |
| pMFCM_S-II:: | 78.14 | 71.95 | 73.51 |
| m:: | 1.40 | 1.20 | 1.20 |
| Para:: Huber $\beta$ | 1 | 1 | 1 |
| KMFCM_Poly:: | 77.87 | 69.70 | 75.03 |
| m:: | 1.80 | 2.80 | 1.40 |
| Para:(Huber;Poly)=$(\beta\|\beta,\theta,d)$ | (1∥3,0.1,2) | (1∥3,1,2) | (1∥3,1,4) |
| KMFCM_RBF:: | 78.03 | 71.53 | 75.56 |
| m:: | 2.20 | 2.40 | 1.40 |
| Para:(Huber;RBF)=$(\beta\|\beta)$ | (1∥3) | (1∥3) | (1∥3) |
| KMFCM_Tanh:: | 78.35 | 71.24 | 75.17 |
| m:: | 1.80 | 2.40 | 1.40 |
| Para:(Huber;Tanh)=$(\beta\|\beta,\theta)$ | (1∥3,1) | (1∥3,1) | (1∥3,0.1) |
| pKMFCM_S-I_Poly:: | 78.53 | 72.56 | 76.03 |
| m:: | 1.20 | 1.20 | 1.40 |
| Para:(Huber;Poly)=$(\beta\|\beta,\theta,d)$ | (1∥1,1,2) | (1∥2,1,2) | (1∥3,1,4) |
| pKMFCM_S-I_RBF:: | 78.70 | 73.02 | 77.95 |
| m:: | 1.40 | 1.20 | 1.40 |
| Para: (Huber;RBF)=$(\beta\|\beta)$ | (1∥1) | (1∥1) | (1∥1) |
| pKMFCM_S-I_Tanh:: | 77.37 | 69.36 | 76.15 |
| m:: | 1.20 | 1.20 | 1.40 |
| Para: (Huber;Tanh)=$(\beta\|\beta,\theta)$ | (1∥1,0.1) | (1∥2,0.1) | (1∥3,1) |
| pKMFCM_S-II_Poly:: | 78.22 | 73.10 | 71.78 |
| m:: | 1.20 | 1.20 | 1.40 |
| Para:(Huber;Poly)=$(\beta\|\beta,\theta,d)$ | (1∥1,1,2) | (1∥1,1,2) | (1∥3,1,4) |
| pKMFCM_S-II_RBF:: | 76.77 | 72.60 | 75.51 |
| m:: | 1.20 | 1.40 | 1.40 |
| Para:(Huber;RBF)=$(\beta\|\beta)$ | (1∥1) | (1∥1) | (1∥1) |
| pKMFCM_S-II_Tanh:: | 77.06 | 69.32 | 74.40 |
| m:: | 1.20 | 1.20 | 1.40 |
| Para: (Huber;Tanh)=$(\beta\|\beta,\theta)$ | (1∥3,0.1) | (1∥2,0.1) | (1∥2,1) |

Table A1_4  weight function  GM

|  | N01 | NoP | U01 |
|---|---|---|---|
| MFCM:: | 58.84 | 57.54 | 73.20 |
| m:: | 3.00 | 3.00 | 1.60 |
| pMFCM_S-I:: | 59.22 | 62.00 | 72.72 |
| m:: | 3.00 | 3.00 | 2.80 |
| pMFCM_S-II:: | 58.89 | 59.31 | 71.47 |
| m:: | 3.00 | 3.00 | 3.00 |
| KMFCM_Poly:: | 78.26 | 71.35 | 75.15 |
| m:: | 3.00 | 2.80 | 1.40 |
| Para: Poly$_{(\beta,\theta,d)}$ | (3,0.1,4) | (3,1,2) | (1,1,2) |
| KMFCM_RBF:: | 75.85 | 72.10 | 74.83 |
| m:: | 2.20 | 2.00 | 1.20 |
| Para: RBF $\beta$ | 3 | 2 | 3 |
| KMFCM_Tanh:: | 75.98 | 72.47 | 74.92 |
| m:: | 2.20 | 2.60 | 1.20 |
| Para: Tanh$_{(\beta,\theta)}$ | (2,1) | (2,1) | (3,0.1) |
| pKMFCM_S-I_Poly:: | 70.86 | 65.69 | 75.91 |
| m:: | 2.80 | 3.00 | 2.20 |
| Para: Poly$_{(\beta,\theta,d)}$ | (1,1,4) | (3,1,2) | (1,1,4) |
| pKMFCM_S-I_RBF:: | 69.55 | 67.65 | 71.57 |
| m:: | 2.20 | 3.00 | 2.60 |
| Para: RBF $\beta$ | 1 | 1 | 1 |
| pKMFCM_S-I_Tanh:: | 69.77 | 65.66 | 70.83 |
| m:: | 2.20 | 3.00 | 2.60 |
| Para: Tanh$_{(\beta,\theta)}$ | (1,0.1) | (1,0.1) | (2,0.1) |
| pKMFCM_S-II_Poly:: | 71.43 | 63.75 | 72.29 |
| m:: | 3.00 | 3.00 | 3.00 |
| Para: Poly$_{(\beta,\theta,d)}$ | (1,1,4) | (2,1,2) | (1,1,2) |
| pKMFCM_S-II_RBF:: | 67.46 | 66.71 | 71.20 |
| m:: | 2.20 | 3.00 | 2.00 |
| Para: RBF $\beta$ | 1 | 1 | 1 |
| pKMFCM_S-II_Tanh:: | 67.60 | 61.15 | 72.74 |
| m:: | 2.20 | 2.80 | 2.80 |
| Para: Tanh$_{(\beta,\theta)}$ | (1,0.1) | (2,0.1) | (2,0.1) |

Table A1_5    weight    function    Welsch

|  | N01 | NoP | U01 |
|---|---|---|---|
| MFCM:: | 75.45 | 69.69 | 74.33 |
| m:: | 1.20 | 1.20 | 1.40 |
| Para:: Welsch $\beta$ | 1 | 3 | 3 |
| pMFCM_S-I:: | 80.19 | 73.55 | 74.73 |
| m:: | 1.40 | 1.20 | 1.60 |
| Para:: Welsch $\beta$ | 1 | 1 | 1 |
| pMFCM_S-II:: | 79.09 | 72.49 | 74.09 |
| m:: | 1.40 | 1.60 | 1.20 |
| Para:: Welsch $\beta$ | 1 | 1 | 1 |
| KMFCM_Poly:: | 77.46 | 70.10 | 74.86 |
| m:: | 1.80 | 2.80 | 1.60 |
| Para: (Welsch;Poly)=$(\beta \| \beta, \theta, d)$ | (1‖2,0.1,2) | (1‖3,0.1,4) | (1‖3,0.1,4) |
| KMFCM_RBF:: | 77.01 | 70.11 | 74.22 |
| m:: | 2.00 | 2.60 | 1.80 |
| Para: (Welsch;RBF)=$(\beta \| \beta)$ | (1‖3) | (1‖3) | (1‖3) |
| KMFCM_Tanh:: | 77.16 | 69.81 | 74.98 |
| m:: | 1.80 | 2.40 | 1.60 |
| Para: (Welsch;Tanh)=$(\beta \| \beta, \theta)$ | (1‖3,1) | (1‖2,1) | (1‖2,1) |
| pKMFCM_S-I_Poly:: | 78.65 | 72.34 | 74.40 |
| m:: | 1.40 | 1.40 | 1.40 |
| Para: (Welsch;Poly)=$(\beta \| \beta, \theta, d)$ | (1‖1,1,4) | (1‖3,1,2) | (1‖2,1,2) |
| pKMFCM_S-I_RBF:: | 78.28 | 72.85 | 77.26 |
| m:: | 1.20 | 1.40 | 1.40 |
| Para: (Welsch;RBF)=$(\beta \| \beta)$ | (1‖1) | (1‖2) | (1‖1) |
| pKMFCM_S-I_Tanh:: | 78.09 | 72.14 | 75.56 |
| m:: | 1.20 | 1.40 | 1.40 |
| Para: (Welsch;Tanh)=$(\beta \| \beta, \theta)$ | (1‖1,1) | (1‖2,0.1) | (1‖1,0.1) |
| pKMFCM_S-II_Poly:: | 78.36 | 72.59 | 73.08 |
| m:: | 1.20 | 1.40 | 1.20 |
| Para: (Welsch;Poly)=$(\beta \| \beta, \theta, d)$ | (1‖1,1,4) | (1‖3,1,2) | (1‖3,1,4) |
| pKMFCM_S-II_RBF:: | 77.44 | 73.52 | 75.47 |
| m:: | 1.20 | 1.40 | 1.20 |
| Para: (Welsch;RBF)=$(\beta \| \beta)$ | (1‖2) | (1‖2) | (1‖1) |
| pKMFCM_S-II_Tanh:: | 77.60 | 72.75 | 74.44 |
| m:: | 1.20 | 1.40 | 1.40 |
| Para: (Welsch;Tanh)=$(\beta \| \beta, \theta)$ | (1‖3,1) | (1‖2,0.1) | (1‖3,0.1) |

Table A1_6   weight   function     Cauchy

|  | N01 | NoP | U01 |
|---|---|---|---|
| MFCM:: | 74.79 | 69.69 | 74.33 |
| m:: | 2.00 | 1.20 | 1.40 |
| Para:: Cauchy $\beta$ | 3 | 3 | 3 |
| pMFCM_S-I:: | 80.23 | 73.21 | 74.83 |
| m:: | 1.40 | 1.20 | 1.60 |
| Para:: Cauchy $\beta$ | 1 | 2 | 1 |
| pMFCM_S-II:: | 79.44 | 72.55 | 74.20 |
| m:: | 1.40 | 1.60 | 1.20 |
| Para:: Cauchy $\beta$ | 1 | 1 | 1 |
| KMFCM_Poly:: | 77.31 | 70.10 | 75.14 |
| m:: | 1.80 | 2.80 | 1.40 |
| Para:(Cauchy;Poly)=$(\beta\|\beta,\theta,d)$ | (1‖2,0.1,2) | (1‖3,0.1,2) | (1‖2,1,4) |
| KMFCM_RBF:: | 76.78 | 70.11 | 74.52 |
| m:: | 2.00 | 2.60 | 1.20 |
| (Cauchy;RBF)=$(\beta\|\beta)$ | (1‖3) | (1‖3) | (1‖1) |
| KMFCM_Tanh:: | 77.26 | 69.97 | 75.11 |
| m:: | 1.80 | 2.40 | 1.60 |
| Para: (Cauchy;Tanh)=$(\beta\|\beta,\theta)$ | (1‖3,1) | (1‖2,1) | (1‖2,1) |
| pKMFCM_S-I_Poly:: | 78.73 | 73.04 | 74.34 |
| m:: | 1.20 | 1.40 | 1.40 |
| Para:(Cauchy;Poly)=$(\beta\|\beta,\theta,d)$ | (1‖1,1,4) | (1‖1,1,2) | (1‖2,1,2) |
| pKMFCM_S-I_RBF:: | 77.99 | 72.24 | 76.45 |
| m:: | 1.40 | 1.40 | 1.20 |
| (Cauchy;RBF)=$(\beta\|\beta)$ | (1‖1) | (1‖1) | (1‖1) |
| pKMFCM_S-I_Tanh:: | 77.97 | 69.46 | 75.41 |
| m:: | 1.20 | 1.20 | 1.40 |
| Para: (Cauchy;Tanh)=$(\beta\|\beta,\theta)$ | (1‖2,0.1) | (1‖3,1) | (1‖2,0.1) |
| pKMFCM_S-II_Poly:: | 78.42 | 72.56 | 72.62 |
| m:: | 1.20 | 1.20 | 1.20 |
| Para:(Cauchy;Poly)=$(\beta\|\beta,\theta,d)$ | (1‖1,1,4) | (1‖2,0.1,2) | (1‖1,1,4) |
| pKMFCM_S-II_RBF:: | 77.18 | 72.77 | 76.65 |
| m:: | 1.20 | 1.40 | 1.20 |
| (Cauchy;RBF)=$(\beta\|\beta)$ | (1‖2) | (1‖1) | (1‖1) |
| pKMFCM_S-II_Tanh:: | 77.20 | 69.43 | 75.30 |
| m:: | 1.20 | 1.20 | 1.20 |
| Para: (Cauchy;Tanh)=$(\beta\|\beta,\theta)$ | (1‖2,0.1) | (1‖2,1) | (1‖1,0.1) |

Table A1_7 weight function Fair

|  | N01 | NoP | U01 |
| --- | --- | --- | --- |
| MFCM:: | 74.84 | 69.56 | 74.31 |
| m:: | 2.00 | 1.20 | 1.20 |
| Para:: Fair $\beta$ | 2 | 3 | 1 |
| pMFCM_S-I:: | 80.45 | 74.70 | 75.08 |
| m:: | 1.40 | 1.80 | 1.80 |
| Para:: Fair $\beta$ | 3 | 1 | 1 |
| pMFCM_S-II:: | 79.20 | 72.95 | 75.04 |
| m:: | 1.40 | 1.20 | 1.20 |
| Para:: Fair $\beta$ | 1 | 3 | 3 |
| KMFCM_Poly:: | 78.03 | 70.60 | 75.79 |
| m:: | 1.20 | 1.20 | 1.60 |
| Para:(Fair;Poly)=$(\beta\|\beta,\theta,d)$ | (1\|\|3,0.1,4) | (1\|\|3,1,2) | (1\|\|3,0.1,2) |
| KMFCM_RBF:: | 77.41 | 69.76 | 74.67 |
| m:: | 2.00 | 2.60 | 1.60 |
| Para: (Fair;RBF)=$(\beta\|\beta)$ | (1\|\|3) | (1\|\|3) | (1\|\|3) |
| KMFCM_Tanh:: | 77.39 | 70.72 | 75.64 |
| m:: | 2.00 | 1.80 | 1.60 |
| Para: (Fair;Tanh)=$(\beta\|\beta,\theta)$ | (1\|\|2,1) | (1\|\|3,0.1) | (1\|\|2,1) |
| pKMFCM_S-I_Poly:: | 79.57 | 72.84 | 76.31 |
| m:: | 1.20 | 1.40 | 1.20 |
| Para:(Fair;Poly)=$(\beta\|\beta,\theta,d)$ | (1\|\|1,1,4) | (1\|\|3,1,2) | (1\|\|3,1,2) |
| pKMFCM_S-I_RBF:: | 78.22 | 72.97 | 76.54 |
| m:: | 1.20 | 1.40 | 1.20 |
| Para: (Fair;RBF)=$(\beta\|\beta)$ | (1\|\|1) | (1\|\|2) | (1\|\|2) |
| pKMFCM_S-I_Tanh:: | 78.20 | 70.37 | 75.55 |
| m:: | 1.20 | 1.40 | 1.40 |
| Para: (Fair;Tanh)=$(\beta\|\beta,\theta)$ | (1\|\|1,0.1) | (1\|\|2,0.1) | (1\|\|2,0.1) |
| pKMFCM_S-II_Poly:: | 78.32 | 72.58 | 74.28 |
| m:: | 1.20 | 1.40 | 1.40 |
| Para:(Fair;Poly)=$(\beta\|\beta,\theta,d)$ | (1\|\|1,1,4) | (1\|\|1,1,2) | (1\|\|2,0.1,4) |
| pKMFCM_S-II_RBF:: | 77.46 | 72.60 | 74.62 |
| m:: | 1.20 | 1.40 | 1.40 |
| Para: (Fair;RBF)=$(\beta\|\beta)$ | (1\|\|1) | (1\|\|2) | (1\|\|1) |
| pKMFCM_S-II_Tanh:: | 77.34 | 71.18 | 74.47 |
| m:: | 1.20 | 1.20 | 1.40 |
| Para: (Fair;Tanh)=$(\beta\|\beta,\theta)$ | (1\|\|1,0.1) | (1\|\|3,1) | (1\|\|2,0.1) |

Note: The detail descriptions of kernal Poly $(\beta,\theta,d)$, RBF $\beta$, Tanh $(\beta,\theta)$, and weights Huber $\beta$, Welsh $\beta$, Cauchy $\beta$, Fair $\beta$ are in Section 5.3.1. The $\beta$s in above Tables are the coefficients of real $\beta$s in respective models.